\documentclass[11pt]{article}

\usepackage[preprint]{acl}

\usepackage{times}
\usepackage{latexsym}

\usepackage[T1]{fontenc}

\usepackage[utf8]{inputenc}

\usepackage{microtype}

\usepackage{inconsolata}

\usepackage{graphicx}

\usepackage{amsmath,amsfonts,bm}

\def\eqref#1{equation~\ref{#1}}

\def\1{\bm{1}}

\DeclareMathAlphabet{\mathsfit}{\encodingdefault}{\sfdefault}{m}{sl}
\SetMathAlphabet{\mathsfit}{bold}{\encodingdefault}{\sfdefault}{bx}{n}

\usepackage{hyperref}
\usepackage{url}
\usepackage{booktabs}
\usepackage{algorithm}
\usepackage{algorithmic}
\usepackage{array}
\usepackage{lipsum}
\usepackage{subcaption}
\usepackage{multirow}
\usepackage{soul}
\usepackage{wrapfig}
\usepackage{enumitem}
\usepackage{graphicx}    
\usepackage{xcolor}      
\usepackage[table]{xcolor}
\usepackage{rotating}
\usepackage{adjustbox}
\usepackage{caption}    
\usepackage{amsmath, amssymb, amsfonts} 
\usepackage[utf8]{inputenc}
\usepackage{threeparttable}
\usepackage{makecell}
\usepackage{csquotes}

\title{Where Matters More Than What: Decoding-aligned KV Cache Compression via Position-aware Pseudo Queries}

\author{Zhenxu Tian$^{1}$\thanks{\; Equal Contribution.}, Yi Su$^1$\footnotemark[1], Juntao Li$^{1}$\thanks{\; Corresponding author.}, Min Zhang$^{1}$ \\  
 $^{1}$School of Computer Science and Technology, Soochow University, China \\
 \texttt{zhenxut@163.com}
 }

\begin{document}
\maketitle
\begin{abstract}
The Key-Value (KV) cache is crucial for efficient Large Language Models (LLMs) inference, but excessively long contexts drastically increase KV cache memory footprint.
Existing KV cache compression methods typically rely on input-side attention patterns within a prompt observation window to estimate token importance during the prefill stage. They fail to preserve critical tokens for future generation since these assessments are not derived from the decoding process.
Intuitively, an effective observation window should mirror the decoding-stage queries to accurately reflect which tokens the generation process will attend to. However, ground-truth decoding queries are inherently unavailable during inference. For constructing pseudo queries to approximate them, we find that positional information plays a more critical role than semantic content. 
Motivated by this insight, we propose decoding-aligned KV cache compression via position-aware pseudo queries (\textbf{DapQ}), a novel and lightweight eviction framework that leverages position-aware pseudo queries to simulate the output tokens, thereby establishing an effective observation window for importance assessment. It aligns closely with the actual generation context and enables precise token eviction.
Extensive evaluations across multiple benchmarks and LLMs demonstrate that DapQ achieves superior performance, particularly under strict memory constraints (e.g., up to nearly lossless performance 99.5\% on NIAH with 3\% KV cache budgets). 
\end{abstract}

\section{Introduction}
Large Language Models \citep{gpt,clip,gemini,deepseek,llama3,qwen3} have achieved significant success across various domains and demonstrated exceptional abilities for processing long-context tasks, such as contextual question answering and document summarization \citep{sumsurvey,deepseek_coder,comprehensive}. A key enabler of efficient inference is the KV cache mechanism, which significantly accelerates autoregressive decoding by reducing the computational complexity of self-attention from $\mathcal{O}(n^2)$ to $\mathcal{O}(n)$. However, with the growth of context length, the memory footprint of KV cache and the high computational overhead increase dramatically, posing a severe obstacle to the efficient deployment and application of LLMs \citep{longbench}.

To tackle these challenges, various methods have been proposed to compress the KV cache, such as token eviction or merging \citep{h2o,keepkv}, quantization \citep{kivi,kvquant}, head or layer-wise sharing \citep{gqa,kvsharer}, low-rank decomposition \citep{get,loki}. Among these, token eviction remains a widely-adopted strategy. Nevertheless, the rapid growth of input length has further intensified the demand for more effective eviction strategies. In response, as implemented in SnapKV\citep{snapkv}, the observation window has proven superior for retaining critical tokens by combining with pooled accumulated attention scores. This approach is further extended by PyramidKV\citep{pyramidkv}, which dynamically allocates layer-wise cache budgets and selects important KV pairs for compression using the window-based attention mechanism. These studies demonstrate the potential of observation windows for effective KV cache compression.

\begin{figure*}
    \centering
    \includegraphics[width=1\linewidth]{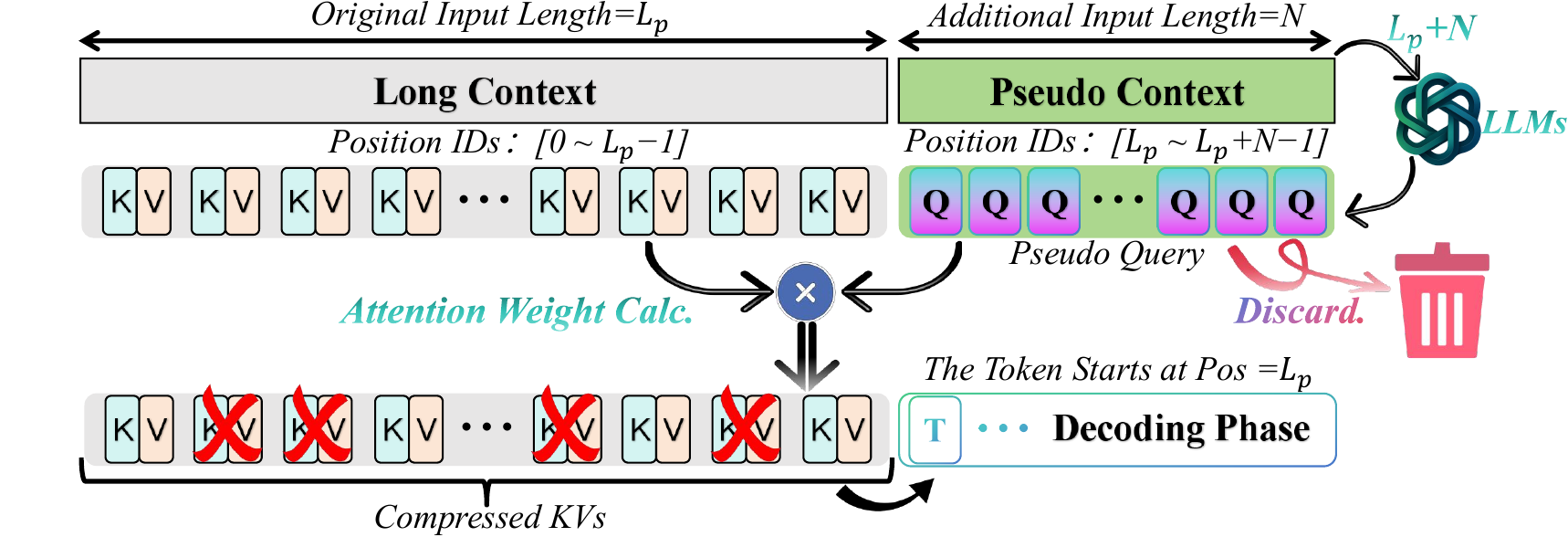}
    \caption{An overview of DapQ. A synthetic pseudo context (length $N$) is appended to the original context (length $L_p$), forming an extended sequence of length $L_p{+}N$. The model processes this sequence during the prefill phase and then obtains pseudo queries for the synthetic tokens, which are endowed with the correct positional encodings of the first $N$ decoding steps. These pseudo queries compute attention scores with all keys from the original prompt, establishing the token importance distribution. The $topK$ tokens are retained in the compressed KV cache, while the others, along with all the synthetic tokens are evicted. Autoregressive decoding then begins from position $L_p$.}
    \label{fig:DapQ}
    \vspace{-10pt}
\end{figure*}

However, the input-centric observation window is inherently misaligned with the dynamic query of actual decoding and relies solely on static prompt-based features, typically the last 16-32 tokens. Consequently, they fail to reflect the importance distribution determined by the output-side generation process, leading to misidentification of the critical tokens for decoding, particularly in complex or noisy contexts. Crucially, ground-truth decoding queries are unavailable during inference, rendering them impractical for directly guiding eviction. To mitigate this, the recent approach LAQ++ \citep{lookahead} attempts to better align the observation window with decoding queries by pre-generating pseudo responses. But its two-stage eviction process introduces a significant memory peak issue that undermines its practical efficiency. Therefore, constructing effective pseudo queries (\texorpdfstring{\(Q_{\text{pseudo}}\)}{Q\_pseudo}) to approximate unavailable future queries without incurring any memory overheads is highly desirable.

Inspired by CaliDrop \citep{calidrop}, where queries at adjacent positions exhibit high similarity, our experiments uncover a pivotal insight: \textbf{positional information plays a more critical role than semantic content in constructing query approximations and determining attention patterns}. This discovery implies that high-quality pseudo queries, capable of reliably assessing the importance distribution of KV cache, can be synthesized based on future positional encodings.

Motivated by this insight, we propose decoding-aligned KV cache compression via position-aware pseudo queries (\textbf{DapQ}), a novel and lightweight KV cache eviction framework that constructs pseudo queries using future positional encodings to accurately simulate the output tokens. These queries collectively serve as an effective observation window for importance scoring that aligns closely with the actual generation context, enabling precise cache eviction. Extensive experiments across multiple benchmarks and different LLMs demonstrate that DapQ achieves superior performance and outperforms existing eviction baselines, particularly under strict memory constraints. 
\label{sec:introduction}
\section{Related Work}


\paragraph{Long-Context LLMs.} The growing demand for LLMs to process long contexts intensifies computational and memory challenges. Prior works address these issues through specialized fine-tuning \citep{longlora} and extending effective context windows via refined positional encodings, such as interpolation and extrapolation \citep{Interpolation,ntk}. To mitigate computational overhead, sparse attention and linear attention have been widely explored \citep{reformer,longformer,linformer}. Beyond traditional Transformer, novel architectures like State-Space Models (SSMs) \citep{longmamba,SSMs} provide linear complexity solutions for processing long sequences. Additionally, memory optimization techniques, such as KV cache compression \citep{streamingllm,snapkv} and memory offloading \citep{hcattention,deepspeed}, have been developed. These multifaceted techniques collectively advance LLMs' capabilities in handling ultra-long sequence tasks.


\begin{table*}[t]
\centering
\resizebox{\textwidth}{!}{%
\begin{tabular}{c >{\raggedright\arraybackslash}p{8.5cm} >{\raggedright\arraybackslash}p{6.5cm} c c}
\toprule
\textbf{Experiment} & \textbf{Content Similarity} & \textbf{Positional Similarity} & \textbf{Post ROPE} & \textbf{Pre ROPE} \\
\midrule
\multirow{2}{*}{SC \& SP} 
& Same\textcolor{gray}{(\enquote{The report discusses the Federal……Airport Improvement Program (AIP). The program})} 
& \multirow{2}{*}{Same\textcolor{gray}{(4424,4425,4426……4453,4454,4455)}} 
& \multirow{2}{*}{\textcolor{green}{1.0000}} & \multirow{2}{*}{\textcolor{green}{1.0000}} \\
\addlinespace
\multirow{2}{*}{DC \& SP} 
& Different\textcolor{gray}{(\enquote{Sorry, I don't know. Sorry, I don't know. Sorry, I don't know. Sorry, I don't know. Sorry, I})} 
& \multirow{2}{*}{Same\textcolor{gray}{(4424,4425,4426……4453,4454,4455)}} 
& \multirow{2}{*}{\textcolor{green}{0.7238}} & \multirow{2}{*}{\textcolor{green}{0.7238}} \\
\addlinespace
\multirow{2}{*}{SC \& DP} 
& Same\textcolor{gray}{(\enquote{The report discusses the Federal……Airport Improvement Program (AIP). The program})} 
& \multirow{2}{*}{Different\textcolor{gray}{(0,1,2……29,30,31)}} 
& \multirow{2}{*}{\textcolor{red}{0.3522}} & \multirow{2}{*}{\textcolor{green}{0.7913}} \\
\addlinespace
\multirow{2}{*}{DC \& DP} 
& Different\textcolor{gray}{(\enquote{Sorry, I don't know. Sorry, I don't know. Sorry, I don't know. Sorry, I don't know. Sorry, I})} 
& \multirow{2}{*}{Different\textcolor{gray}{(0,1,2……29,30,31)}} 
& \multirow{2}{*}{\textcolor{red}{0.3267}} & \multirow{2}{*}{\textcolor{green}{0.7434}} \\
\bottomrule
\end{tabular}%
}
\vspace{-8pt}
\caption{Query similarity comparison under different content and position conditions. Post ROPE denotes similarity after ROPE has been applied to query vectors. Pre ROPE indicates similarity measured before ROPE application.}
\label{table:query_similarity}
\vspace{-8pt}
\end{table*}

\begin{figure*}
    \centering
    \begin{subfigure}[t]{0.47\textwidth}
        \centering
        \resizebox{\textwidth}{!}{\includegraphics{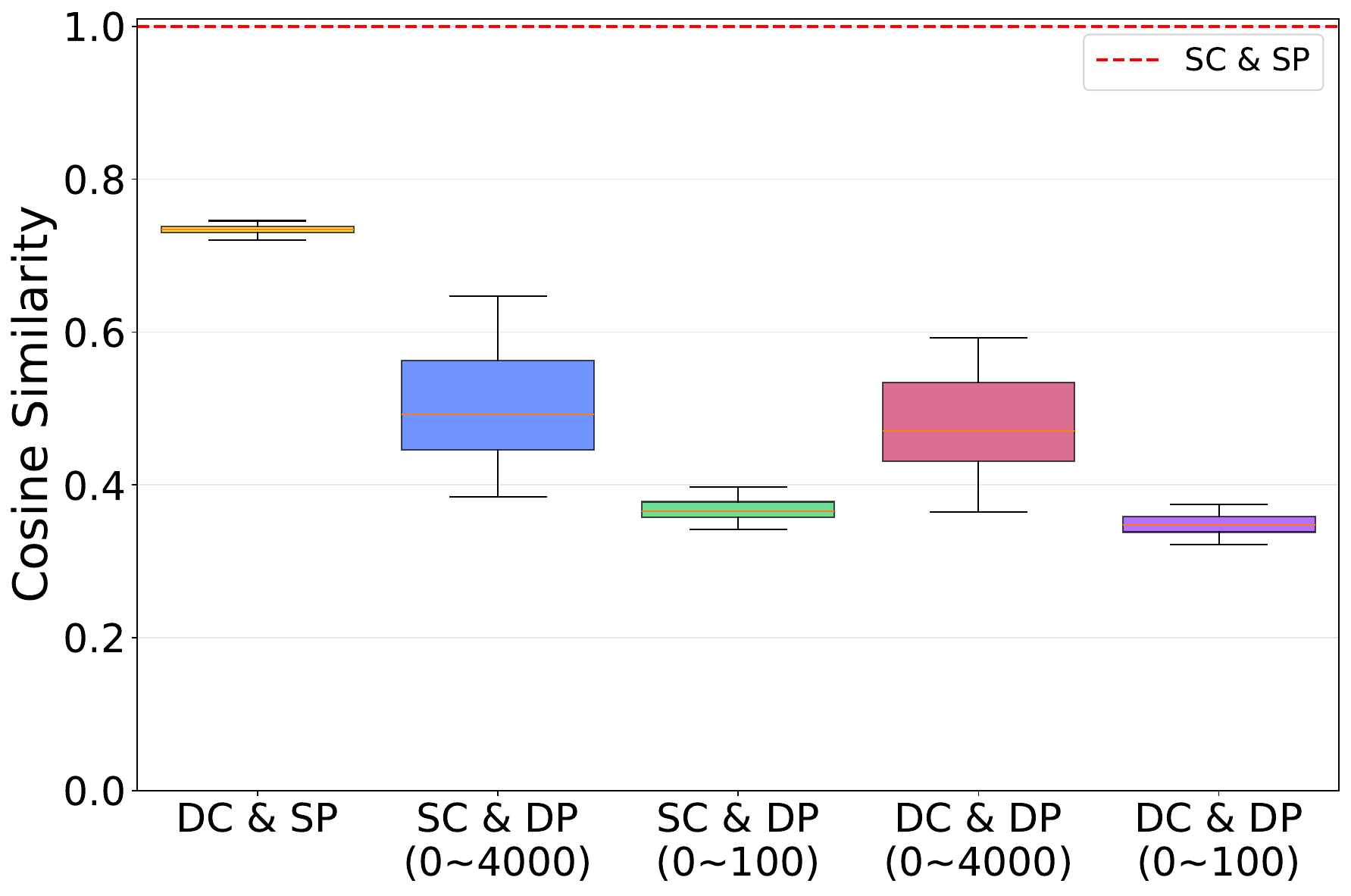}}
        \caption{}  
        \label{fig:query_similarity_boxplot}
    \end{subfigure}
    \hspace{0.5cm}  
    \begin{subfigure}[t]{0.47\textwidth}
        \centering
        \resizebox{\textwidth}{!}{\includegraphics{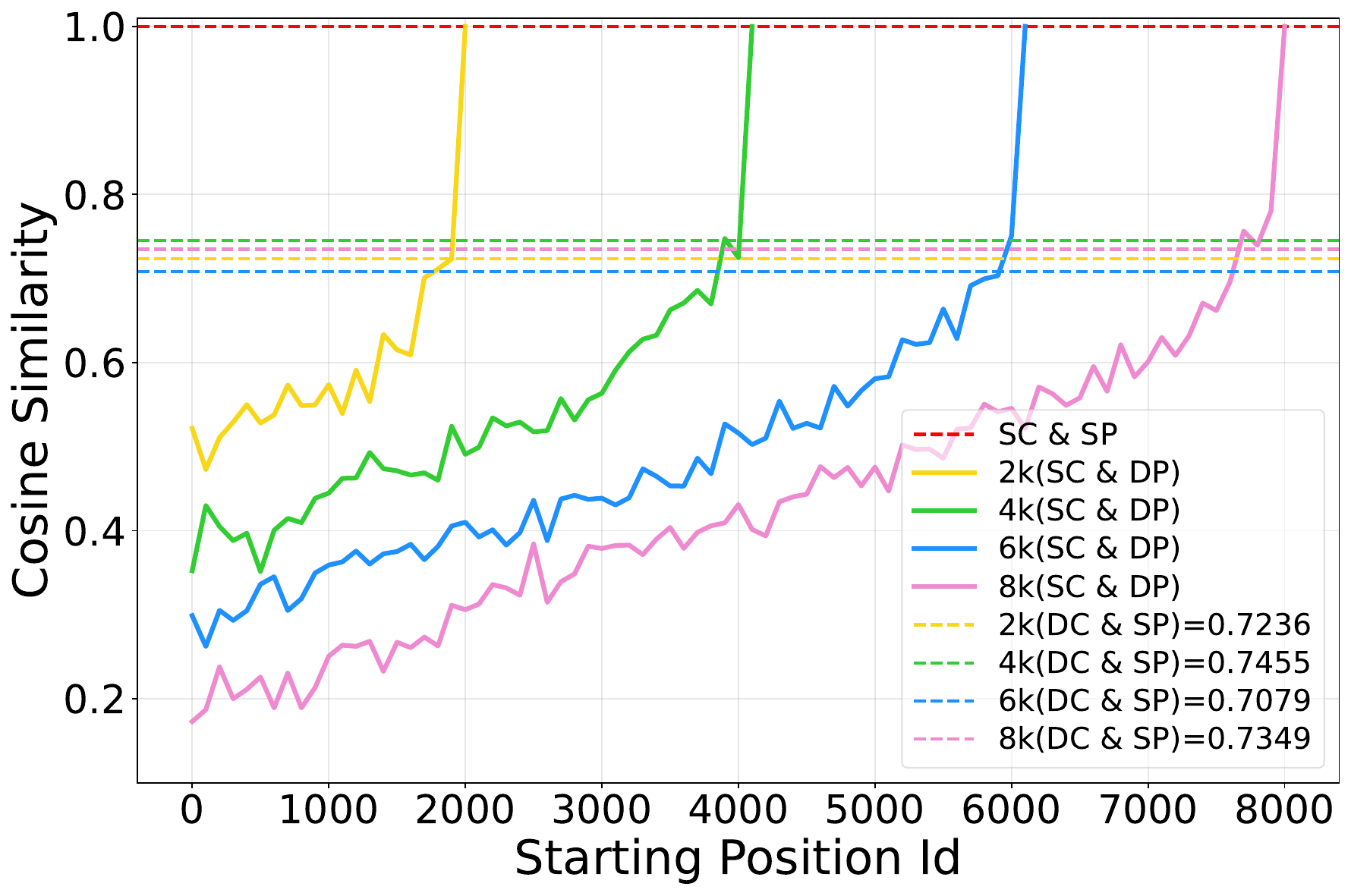}}
        \caption{}
        \label{fig:query_similarity_curve}
    \end{subfigure}
    \vspace{-7pt}
    \caption{Analysis of Positional Dominance and Offset Sensitivity in Query Similarity. We set the pseudo queries of fixed length 32. (a) Boxplot of query similarity distributions for a 4k context under different content and position conditions, each aggregated from 100 independent trials. DC: pseudo-queries content is constructed by randomly sampling 32 tokens from the model's vocabulary; DP: pseudo-queries positions are assigned by randomly sampling a consecutive span of 32 index positions from the context length range [0, m] (e.g., [0, 4000]). (b) Query similarity curves over offset positions for contexts of lengths 2k, 4k, 6k, and 8k. The x-axis denotes the starting position assigned to pseudo queries (e.g., an x-axis value of 3500 corresponds to position IDs $3500\sim3531$).}

    \vspace{-14pt}
\end{figure*}

\paragraph{KV Cache Compression.} KV cache compression is crucial for enhancing the inference efficiency and deployability of LLMs, particularly in resource-constrained scenarios. Various methods have been developed to reduce KV cache memory footprint. Token eviction strategies aim to retain only the most important tokens based on metrics like attention scores \citep{snapkv,h2o}, positional heuristics \citep{streamingllm}, special tokens \citep{gemodel,sepllm}, or norm-based criteria \citep{simple}. Quantization \citep{kivi,kvquant} reduces memory by storing less important KV pairs with lower precision, some approaches even achieving sub-2-bit quantization via token-aware and channel-aware techniques. Sharing-based approaches deliver memory savings and accelerate inference through head-wise sharing \citep{mqa,gqa}, inter-layer sharing \citep{sun2024you,wu2024layer,brandon2024reducing}, or prefix sharing across sequences \citep{hydragen,relayattention}. Low-rank decomposition \citep{gear,palu} projects KV cache into lower-dimensional spaces to exploit inherent redundancy, as demonstrated by the Multi-Head Latent Attention \citep{deepseekv3} of DeepSeek, which reduces cache size through low-rank compression and decoupled RoPE while preserving model performance. KV merging \citep{keepkv,homogeneous} employs attention-pattern similarity or reparameterization to merge similar semantic information, achieving effective compression with minimal performance loss.

\section{Observation}


Given the discussion in Section \ref{sec:introduction}, constructing pseudo queries to accurately approximate the unavailable ground-truth decoding queries becomes crucial. Building upon CaliDrop's \citep{calidrop} insight that queries at adjacent positions exhibit high similarity, we hypothesize that this similarity is strongly correlated with positional information rather than semantic content. This prompts us to investigate whether positional information alone can effectively approximate future decoding queries without relying on true decoding content. See Appendix \ref{Preliminary Experiment Details} for details of preliminary experiments.


\subsection{Position Drives Query Representation}
As detailed in Table \ref{table:query_similarity},We compare cosine similarities between ground-truth decoding queries and pseudo queries across four conditions: SC (Same Content), DC (Different Content), SP (Same Position), and DP (Different Position). Specifically, pseudo queries assigned correct future positional IDs but composed of completely irrelevant or nonsensical content (\textcolor{green}{DC\&SP}), exhibit strong cosine similarity (\textcolor{green}{0.7238}) to the actual target decoding queries. Conversely, queries with the identical semantic content but incorrect positional IDs (\textcolor{red}{SC\&DP} vs \textcolor{green}{SC\&SP}) fail to accurately approximate the target queries (\textcolor{red}{0.3522} vs \textcolor{green}{1.0000}). Notably, the comparison between DC\&SP and SC\&DP highlights that maintaining correct positional alignment achieves 2.1× higher similarity than maintaining correct content alone. The stark contrast underscores that the semantic content of queries plays a secondary role compared to positional encodings in constructing query approximations. The consistently high similarity under Pre-ROPE conditions (\textcolor{green}{0.7238} to \textcolor{green}{0.7913}) confirms that the model's underlying processing does not rely heavily on semantic content to distinguish between these query vectors, thereby underscoring that the dramatic disparity observed in Post-ROPE scores is almost attributable to the positional information. A large-scale statistical analysis (Figure \ref{fig:query_similarity_boxplot}) confirms the pervasiveness and consistency of the above phenomenon.

\subsection{Precise Positional Alignment is Necessary}
Building upon the dominance of positional information, we further quantify how the positional alignment of pseudo queries affects their similarities to true decoding queries. As shown in Figure \ref{fig:query_similarity_curve}, we fix the pseudo-query semantic content to match the true output and systematically vary their assigned positional IDs. \textbf{Results reveal a monotonic decay in the query similarity as the absolute offset increases between the assigned position and the correct position}. This decay phenomenon is consistently observed across diverse context lengths (2k to 8k), which is more pronounced in longer context scenarios. The strong sensitivity to positional misalignment underscores the critical dependence of query approximations on precise positional information. Consequently, accurately simulating decoding queries requires precise alignment with the future generation positional IDs.

\begin{figure}[htbp]
    \centering
    \includegraphics[width=\linewidth]{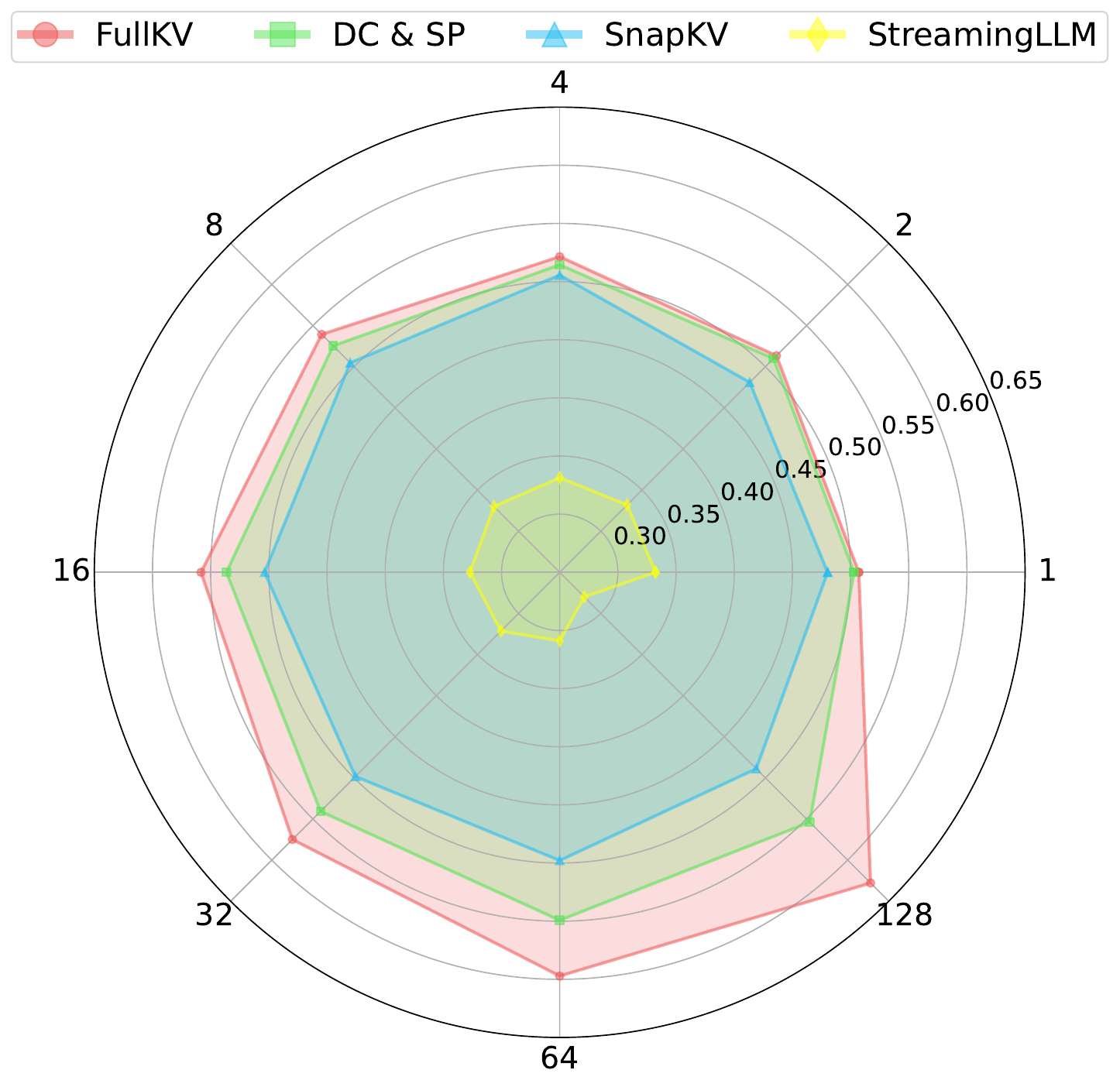} 
    \vspace{-18pt}
    \caption{Recall Performance of different methods across various Window Sizes.} 
    \label{fig:recall_radar} 
    \vspace{-18pt}
\end{figure}

\subsection{Position-Based Queries Enable More Accurate Token Eviction}
The critical question is whether higher query similarity translates into more accurate token eviction. To quantify this, we evaluate the recall of eviction strategies, which measures its ability to retain the tokens that are most important for the actual generation. Following the methodology of prior work \citep{lookahead}, the recall rate of the selected KV cache is defined as the proportion of indices selected by the observation window that overlap with those selected by all response tokens from the model. We define the recall metric as follows:

Let $R$ be the set of all tokens in the ground-truth response, and let $K$ be the full key cache from the prefill stage. The gold standard set of indices $M_{\text{gold}}$ for a given budget $B$ is determined by accumulated attention scores from all true response queries:
\begin{equation}
M_{gold} = \underset{{i \in [0, N)}}{TopK}\left( \sum_{j \in R} \text{Attention}(q_j, K) \right),
\end{equation}
where $N$ is the number of all prefill tokens, and $TopK$ returns the indices of the tokens with the highest accumulated scores. The predicted set of indices $M_{\text{pred}}$ is defined analogously to $M_{\text{gold}}$, but computed using only the queries in a candidate observation window $W$:
\begin{equation}
M_{pred} = \underset{{i \in [0, N)}}{TopK}\left( \sum_{j \in W} \text{Attention}(q_j, K) \right).
\end{equation}

The recall is the proportion of the gold-standard tokens that are correctly retained:

\begin{equation}
Recall_{W} = \frac{| M_{\text{gold}} \cap M_{pred} |}{| M_{gold} |}.
\end{equation}

We evaluate this recall metric on the GovReport dataset for different observation windows. As shown in Figure \ref{fig:recall_radar}, the window composed of pseudo queries with randomized content but correct future positions achieves significantly better recall than baselines like SnapKV and StreamingLLM. Notably, it maintains high recall even as the window size is reduced to 32 or 16, showing the high effectiveness and accuracy of position-based estimation. This demonstrates that a small set of pseudo queries, informed solely by precise positional forecasting, provides a highly effective basis for importance assessment, enabling accurate eviction even under extreme window constraints.

In summary, our experiments converge on a pivotal insight: the representation of a query vector is dominated by its positional encoding, with semantic content playing a secondary role. From the perspective of query-key interactions, this further implies that the attention pattern relies heavily on positional information to establish importance distribution, while exhibiting considerable robustness to variations in semantic content. This leads to a profound practical implication: high-fidelity decoding pseudo queries can be synthesized from positional encodings, entirely bypassing the computationally expensive and memory-intensive process of token generation. This position-aware query approximation forms the foundation of our method.


\section{Method}

Motivated by the pivotal insight that positional information dominates query representations, we propose DapQ \footnote{\url{https://github.com/tianzhenxu/DapQ}}(as illustrated in Figure \ref{fig:DapQ}), a novel KV cache compression framework that accurately simulates decoding-stage contextual positioning during the prefill phase. DapQ synthesizes a decoding-aligned observation window, composed of pseudo queries endowed with future positional encodings, which mirrors the dynamic context of the actual decoding process. This precisely assesses token importance, enabling accurate token eviction without altering the intended timeline.

\subsection{Construct Decoding-Aligned \texorpdfstring{\(Q_{\text{pseudo}}\)}{Q\_pseudo}}
The core of DapQ is to simulate the dynamic positional query of the decoding phase. For a prompt sequence of length \(L_p\), we append a set of \(N\) artificially constructed tokens, denoted \(\mathbf{T}_{\text{pseudo}}\), to form an extended input sequence. \(\mathbf{T}_{\text{pseudo}}\) can be constructed or arbitrarily chosen from the existing context (e.g., uniformly sampled or prefix-suffix concatenation), as their semantic content is secondary to the positional assignment. The crucial operation is to assign \(\mathbf{T}_{\text{pseudo}}\) the correct positional indices that they would occupy as the first \(N\) tokens generated by the model, rather than arbitrarily:
\begin{equation*}
\text{Positions}(\mathbf{T}_{\text{pseudo}}) = [L_p, L_p+1, ..., L_p+N-1].
\end{equation*}
This yields an input sequence with length \(L_{\text{total}} = L_p + N\). The model processes this extended sequence during the prefill phase, computing KV cache for \(L_{\text{p}}\) prompt tokens. The primary purpose of this step is to obtain pseudo queries (\(Q_{\text{pseudo}}\)) of these  \(\mathbf{T}_{\text{pseudo}}\), which are endowed with correct positional encodings for the start of decoding phase.

\subsection{Importance Assessment and Eviction}
We leverage the \(Q_{\text{pseudo}}\) to assess the importance of all Keys derived from the original prompt. The importance score for the \(j\)-th (\(j \in [0, L_p - 1]\)) prompt token is computed by aggregating its attention scores from each pseudo query \(q_i \in Q_{\text{pseudo}}\):
\begin{equation}
S(j) = \sum_{i=L_p}^{L_p+N-1} \text{Attention}(q_i, k_j).
\end{equation}
The \(TopK\) tokens with the highest scores \(S(j)\) are retained:
\begin{equation}
M_{\text{retain}} = \underset{j \in [0, L_p - 1]}{TopK}\left( S(j) \right).
\end{equation}
The KV cache is pruned, discarding all key-value pairs not in \(M_{\text{retain}}\). \textbf{Crucially, the entire synthetic segment \(\mathbf{T}_{\text{pseudo}}\) is discarded immediately after performing the importance scoring.} Autoregressive decoding phase then begins from position \(L_p\), utilizing only the compressed cache of size \(K\). This ensures the model's generation remains consistent with the intended timeline.

\section{Experiments}

\begin{table*}[t]
\centering
\small
\resizebox{\textwidth}{!}{%
\begin{tabular}{>{\bfseries}c >{\raggedright}p{1.2cm} *{14}{c}}
\toprule
&  & \multicolumn{2}{c}{\textbf{Single-Document QA}} & \multicolumn{2}{c}{\textbf{Multi-Document QA}} & \multicolumn{2}{c}{\textbf{Summarization}} & \multicolumn{3}{c}{\textbf{Few-shot Learning}} & \multicolumn{2}{c}{\textbf{Synthetic}} & \multicolumn{2}{c}{\textbf{Code}} &  \\
\cmidrule(lr){3-4}\cmidrule(lr){5-6}\cmidrule(lr){7-8}\cmidrule(lr){9-11}\cmidrule(lr){12-13}\cmidrule(lr){14-15}
& \textbf{Methods} & \rotatebox[origin=c]{30}{\textbf{Qasper}} & \rotatebox[origin=c]{30}{\textbf{MF-en}} & \rotatebox[origin=c]{30}{\textbf{HotpotQA}} & \rotatebox[origin=c]{30}{\textbf{2WikiMQA}} & \rotatebox[origin=c]{30}{\textbf{GovReport}} & \rotatebox[origin=c]{30}{\textbf{MultiNews}} & \rotatebox[origin=c]{30}{\textbf{TREC}} & \rotatebox[origin=c]{30}{\textbf{TriviaQA}} & \rotatebox[origin=c]{30}{\textbf{SAMSum}} & \rotatebox[origin=c]{30}{\textbf{PCount}} & \rotatebox[origin=c]{30}{\textbf{PRe}} & \rotatebox[origin=c]{30}{\textbf{Lcc}} & \rotatebox[origin=c]{30}{\textbf{RB-P}} & \textbf{Avg.} \\
\midrule
& FullKV & 37.72 & 40.64 & 50.17 & 34.88 & 31.03 & 25.64 & 70.00 & 89.85 & 40.55 & 13.30 & 83.67 & 58.96 & 52.71 & 48.39 \\
& \multicolumn{15}{c}{\cellcolor{gray!20}\textbf{KV Cache Size = 256}} \\
\multirow{13}{*}{\rotatebox{90}{Llama3-8B-Instruct}} & H2O & 28.11 & 36.63 & 48.62 & 31.50 & 21.87 & 21.44 & 45.67 & 89.49 & 38.28 & 12.11 & 83.67 & 61.49 & 53.36 & 44.02 \\
& PyramidKV & 30.88 & 38.11 & 50.20 & 33.88 & \textbf{22.54} & 21.84 & 60.00 & 89.26 & 37.07 & \textbf{12.78} & 83.67 & 61.34 & 52.51 & 45.70 \\
& SnapKV & 30.84 & \textbf{38.39} & 49.75 & 33.80 & 22.18 & 21.53 & 57.00 & 89.65 & 36.97 & 12.11 & \textbf{84.00} & 61.78 & 54.92 & 45.61 \\
& DapQ & \textbf{32.55} & 38.18 & \textbf{50.67} & \textbf{34.35} & 22.25 & \textbf{21.89} & \textbf{60.67} & \textbf{90.48} & \textbf{38.34} & 11.78 & 83.67 & \textbf{62.78} & \textbf{55.64} & \textbf{46.40} \\
& \multicolumn{15}{c}{\cellcolor{gray!20}\textbf{KV Cache Size = 128}} \\
& H2O & 25.95 & 36.25 & 48.65 & 31.90 & \textbf{20.79} & 20.30 & 40.00 & 87.29 & 36.25 & 12.33 & \textbf{83.67} & 59.81 & 53.14 & 42.79 \\
& PyramidKV & 28.80 & \textbf{38.29} & 49.52 & 31.60 & 20.67 & 20.55 & 49.00 & 87.68 & 36.73 & \textbf{12.44} & 82.00 & 60.36 & 52.03 & 43.82 \\
& SnapKV & \textbf{29.52} & 37.80 & 49.36 & 32.40 & 19.87 & 20.08 & 47.67 & 87.82 & 35.63 & 11.44 & 82.33 & 61.49 & 52.40 & 43.68 \\
& DapQ & 28.76 & 37.24 & \textbf{50.04} & \textbf{33.59} & 20.47 & \textbf{20.63} & \textbf{50.00} & \textbf{90.06} & \textbf{36.87} & 12.11 & 81.67 & \textbf{61.81} & \textbf{53.92} & \textbf{44.40} \\
& \multicolumn{15}{c}{\cellcolor{gray!20}\textbf{KV Cache Size = 64}} \\
& H2O & 24.02 & 30.83 & 48.27 & 31.70 & \textbf{19.37} & \textbf{19.14} & 37.33 & 86.27 & 35.18 & 7.72 & \textbf{82.33} & 59.20 & \textbf{51.10} & 40.96 \\
& PyramidKV & 22.04 & 31.80 & 47.01 & 31.54 & 15.70 & 16.34 & 39.00 & 76.80 & 32.31 & 10.33 & 79.67 & 55.19 & 47.90 & 38.90 \\
& SnapKV & 25.06 & 32.92 & 47.16 & 31.71 & 16.85 & 17.09 & \textbf{40.67} & 86.02 & 33.99 & 11.78 & 78.00 & 57.95 & 50.91 & 40.78 \\
& DapQ & \textbf{25.99} & \textbf{37.36} & \textbf{49.11} & \textbf{32.88} & 18.46 & 18.70 & 38.67 & \textbf{87.38} & \textbf{35.30} & \textbf{11.89} & 77.67 & \textbf{60.19} & 49.90 & \textbf{41.81} \\
\midrule
& FullKV & 36.87 & 53.67 & 57.67 & 44.73 & 33.39 & 23.69 & 71.67 & 91.79 & 42.07 & 11.98 & 86.67 & 70.64 & 59.26 & 52.60 \\
& \multicolumn{15}{c}{\cellcolor{gray!20}\textbf{KV Cache Size = 256}} \\
\multirow{12}{*}{\rotatebox{90}{Qwen3-8B}} & H2O & 27.80 & 44.92 & 51.37 & 40.50 & 22.38 & 18.26 & 46.33 & 90.04 & 38.98 & 12.33 & 86.67 & 66.11 & 53.93 & 46.12 \\
& PyramidKV & 30.41 & 47.81 & 51.00 & 41.37 & 22.36 & 17.41 & 61.67 & 90.96 & 37.65 & 13.00 & 86.33 & 67.01 & 50.11 & 47.47 \\
& SnapKV & \textbf{32.40} & 49.29 & 54.38 & 41.40 & 23.79 & 18.99 & \textbf{63.00} & 91.07 & 38.57 & \textbf{14.67} & 86.67 & \textbf{67.99} & 53.77 & 48.92 \\
& DapQ & 32.14 & \textbf{50.78} & \textbf{54.79} & \textbf{44.47} & \textbf{24.16} & \textbf{19.01} & 62.67 & \textbf{91.15} & \textbf{39.61} & 14.17 & \textbf{86.67} & 67.20 & \textbf{53.83} & \textbf{49.28} \\
& \multicolumn{15}{c}{\cellcolor{gray!20}\textbf{KV Cache Size = 128}} \\
& H2O & 26.60 & 41.37 & 48.10 & 39.85 & 20.83 & 17.27 & 41.33 & 90.21 & 38.16 & \textbf{11.72} & \textbf{86.67} & 65.22 & 52.77 & 44.22 \\
& PyramidKV & 26.22 & 40.48 & 48.41 & 39.46 & 18.99 & 15.10 & 48.33 & 89.31 & 36.92 & 9.67 & 86.67 & 60.82 & 48.78 & 43.78 \\
& SnapKV & \textbf{29.41} & 46.43 & 51.20 & 41.66 & 20.37 & 16.64 & 51.33 & \textbf{91.07} & 37.37 & 11.00 & 86.67 & 65.36 & 51.65 & 46.17 \\
& DapQ & 29.10 & \textbf{47.15} & \textbf{53.84} & \textbf{43.00} & \textbf{21.11} & \textbf{17.27} & \textbf{54.00} & 90.45 & \textbf{38.50} & 11.67 & 86.00 & \textbf{66.29} & \textbf{52.03} & \textbf{46.95} \\
& \multicolumn{15}{c}{\cellcolor{gray!20}\textbf{KV Cache Size = 64}} \\
& H2O & 25.55 & 38.94 & 46.66 & 39.27 & \textbf{18.55} & \textbf{15.23} & 39.00 & 88.13 & 35.98 & 9.67 & \textbf{86.67} & 59.48 & \textbf{48.95} & 42.47 \\
& PyramidKV & 25.32 & 40.44 & 46.61 & 39.20 & 16.25 & 12.93 & \textbf{44.67} & 88.27 & 34.63 & 11.33 & 83.67 & 59.73 & 46.48 & 42.27 \\
& SnapKV & 25.09 & 39.89 & 46.58 & 39.38 & 15.28 & 12.38 & 42.67 & 87.93 & 35.12 & 11.33 & 84.33 & 57.96 & 46.49 & 41.88 \\
& DapQ & \textbf{25.78} & \textbf{43.17} & \textbf{49.84} & \textbf{41.28} & 17.16 & 13.95 & 43.00 & \textbf{88.97} & \textbf{36.00} & \textbf{12.67} & 83.00 & \textbf{60.31} & 46.90 & \textbf{43.23} \\
\bottomrule
\end{tabular}
}
\vspace{-9pt}
\caption{Main Results on LongBench: performance comparison of different KV cache compression methods}
\vspace{-12pt}
\label{table:longbench}
\end{table*}

\begin{table}[t]
    \centering
    \small 
    \setlength{\tabcolsep}{3pt}
    \renewcommand{\arraystretch}{0.85}
    \resizebox{\columnwidth}{!}{
        \begin{tabular}{lcccccc}
            \toprule
            \multirow{2}{*}{\textbf{Method}} & \multicolumn{6}{c}{\textbf{Batch Size}} \\
            \cmidrule(lr){2-7}
            & 1 & 10 & 20 & 30 & 40 & 50 \\
            \midrule
            FullKV     & 11.59 & 26.43 & 25.60 & 26.59 & OOM   & OOM   \\
            LaCache    & 11.44 & 35.77 & 40.64 & 42.49 & OOM   & OOM   \\
            SLM        & 10.81 & 34.49 & 38.21 & 39.25 & 39.98 & 40.46 \\
            H2O        & 10.56 & 34.34 & 37.71 & 39.02 & 39.62 & 40.01 \\
            PyramidKV  & 10.54 & 34.12 & 38.00 & 39.03 & 39.86 & 40.11 \\
            SnapKV     & 10.77 & 34.22 & 38.09 & 39.10 & 39.77 & 40.23 \\
            DapQ       & 10.68 & 34.16 & 37.99 & 38.97 & 39.73 & 40.12 \\
            \bottomrule
        \end{tabular}
    }
    \vspace{-10pt}
    \caption{Comparison of throughput (tokens/s) with different batch sizes.}
    \vspace{-12pt}
    \label{tab:throughput_comparison}
\end{table}

    
\begin{table}[t]
    \centering
    \small
    \setlength{\tabcolsep}{3pt}
    \renewcommand{\arraystretch}{0.85}
    \resizebox{\columnwidth}{!}{
        \begin{tabular}{lccccc}
            \toprule
            \multirow{2}{*}{\textbf{Method}} & \multicolumn{5}{c}{\textbf{Context Length}} \\
            \cmidrule(lr){2-6}
            & 8K & 16K & 32K & 64K & 128K \\
            \midrule
            FullKV     & 1.1106 & 2.5607 & 6.5718 & 18.9441 & 60.8399 \\
            LaCache    & 1.1283 & 2.5921 & 6.6356 & 19.0519 & 61.5180 \\
            SLM        & 1.1236 & 2.5801 & 6.6008 & 18.9853 & 61.0339 \\
            H2O        & 1.1348 & 2.6046 & 6.6352 & 19.0379 & 61.5029 \\
            PyramidKV  & 1.1253 & 2.5958 & 6.6337 & 19.0462 & 61.5345 \\
            SnapKV     & 1.1278 & 2.5909 & 6.6242 & 19.0318 & 61.5017 \\
            DapQ       & 1.1298 & 2.5974 & 6.6289 & 19.0423 & 61.5097 \\
            \bottomrule
        \end{tabular}
    }
    \vspace{-10pt}
    \caption{Comparison of Time-to-First-Token (TTFT (s)) with different context lengths.}
    \vspace{-22pt}
    \label{tab:ttft_comparison}
\end{table}

\subsection{Settings}
\textbf{Models and Benchmarks.}
To evaluate the applicability and generalization of DapQ in various models, we conduct experiments on LLaMA-3-8B-Instruct, LLaMA-3.1-8B-Instruct \citep{llama3}, Qwen2.5-7B-Instruct \citep{qwen2}, and Qwen3-8B \citep{qwen3}. To ensure a more comprehensive and robust assessment, we use five benchmarks: LongBench \citep{longbench}, LongBenchV2 \citep{longbenchv2}, Ruler \citep{ruler}, HELMET \citep{helmet}, and Needle-in-a-Haystack \citep{needle}, each designed to assess distinct aspects of long-context inference, thereby forming a solid foundation for validating DapQ's performance across diverse scenarios. Due to space limitations, complete experiment results and details are presented in Appendix \ref{Complete Experiment Results}.

\textbf{Baselines.}
To comprehensively validate the performance of DapQ, we select six representative KV cache compression methods as baselines: \textbf{FullKV} caches all keys and values for every token, which is the standard approach for KV Cache in transformer-based models; \textbf{SnapKV} \citep{snapkv} captures attention signals from an observation window and uses pooling-based clustering to select important KV pairs for compression; \textbf{PyramidKV} \citep{pyramidkv} leverages cross-layer attention distribution characteristics to dynamically allocate different KV cache budgets and selects important KV pairs for compression; \textbf{H2O} \citep{h2o} identifies Heavy Hitter (H2) tokens based on cumulative attention scores and dynamically balances the retention of recent and H2 tokens to compress KV cache; \textbf{StreamingLLM (SLM)} \citep{streamingllm} identifies the attention sink and dynamically balances the retention of recent and initial tokens to compress KV cache; \textbf{LaCache} \citep{lacache} adopts a ladder-shaped pattern in the prefilling stage to retain KV of early tokens in shallow layers and gradually shift to later tokens in deeper layers. \textbf{Note:} To ensure rigor and consistency, compression is performed solely during the prefill stage.

\textbf{Implementation Details.}
For all methods, we set the observation window size to 32 unless otherwise specified (e.g., LaCache use its default settings). In DapQ, pseudo queries are constructed by concatenating a small number of tokens from the beginning and the end of the input sequence(e.g., the first 4 and last 28 tokens, the first 2 and last 30 tokens). This design is motivated by two key considerations: the beginning tokens, often high-frequency special tokens (e.g.,\texttt{<|begin\_of\_text|>}), possess stable and generalizable embeddings due to their extensive exposure during training; the ending tokens carry the most recent context, making their semantic state highly relevant to the imminent decoding step. This finding is further supported by \citet{prefix_suffix}. We also validate it through experiments in Figure \ref{fig:query_quality}, where concatenating prefix and suffix tokens consistently yields superior performance compared to using random or intermediate consecutive tokens from the input as query contents.

\subsection{Improvement on Accuracy}
Consistent accuracy gains across benchmarks:
\textbf{LongBench Results.} As shown in Table \ref{table:longbench}, DapQ consistently outperforms all baselines across different models and cache budgets. The advantage is particularly pronounced under aggressive compression (e.g., budget=64), where DapQ shows a robust ability to retain critical information (e.g., preserving the long-range contextual dependencies especially on complex information integration tasks like HotpotQA and 2WikiMQA) and mitigate high-compression performance degradation. 
\textbf{LongBenchV2 Results.}
Under a 64 cache budget, DapQ achieves 29.26\% accuracy in the category of \enquote{Hard}, marking a +6.75\% absolute improvement over SnapKV (22.51\%) on LLaMA3-8B-Instruct.
\textbf{Ruler Results.}
Table \ref{table:ruler_llama3} shows that DapQ attains a notable 59.6\% accuracy on the challenging S-NIAH-3 task with a cache budget of 512, substantially outperforming SnapKV (1.4\%) and H2O (2.4\%). 
\textbf{HELMET Results.}
As reported in Table \ref{table:helmet}, DapQ achieves an average score of 48.10 on Qwen2.5-7B-Instruct with a low cache budget of 512, surpassing strong baselines SnapKV (43.74), H2O (40.36), and PyramidKV (42.49). 
\textbf{Needle-in-a-Haystack Results.}
As shown in Figure \ref{fig:needle} and Table \ref{table:needle}, a striking example is on LLaMA3-8B with a cache size of 256: DapQ achieves 99.5\% accuracy, closely approaching full-cache performance. This exceptional performance shows DapQ effectively simulates the decoding-stage positional context via prospectively encoded pseudo queries, enabling precise identification and retention of the key “needles” amidst a vast “haystack” of tokens.

Across diverse benchmarks, DapQ demonstrates the highest average score across nearly all budget settings on different models and especially delivers strong performance gains under strict cache constraints. These results underscores its effectiveness, robustness and generalizability in identifying and preserving critical contextual information.



\subsection{Analysis on Efficiency}
We conduct a comprehensive efficiency evaluation of DapQ using Llama-3.1-8B-Instruct. We first focus on memory usage and throughput across different batch sizes (Input 8k, Output 150 tokens, Budget=256). As shown in Table \ref{tab:throughput_comparison} and Table \ref{tab:memory_comparison}, DapQ maintains robust performance, exhibiting memory and throughput highly on par with other compression methods. Furthermore, to assess the algorithmic overhead introduced during the prefill stage, we measure the Time-to-First-Token (TTFT) across varying sequence lengths from 8K to 128K. Table \ref{tab:ttft_comparison} indicates that the latency of DapQ is nearly identical to that of SnapKV. The results confirms that the additional prefill-stage overhead is almost negligible. Overall, DapQ achieves a balanced performance between long-context understanding capability and inference efficiency, ensuring its practicality for real-world long-context applications.

\section{Analysis}
\begin{figure*}
    \centering
    \begin{subfigure}[t]{0.32\textwidth}  
        \centering
        \resizebox{\textwidth}{!}{\includegraphics{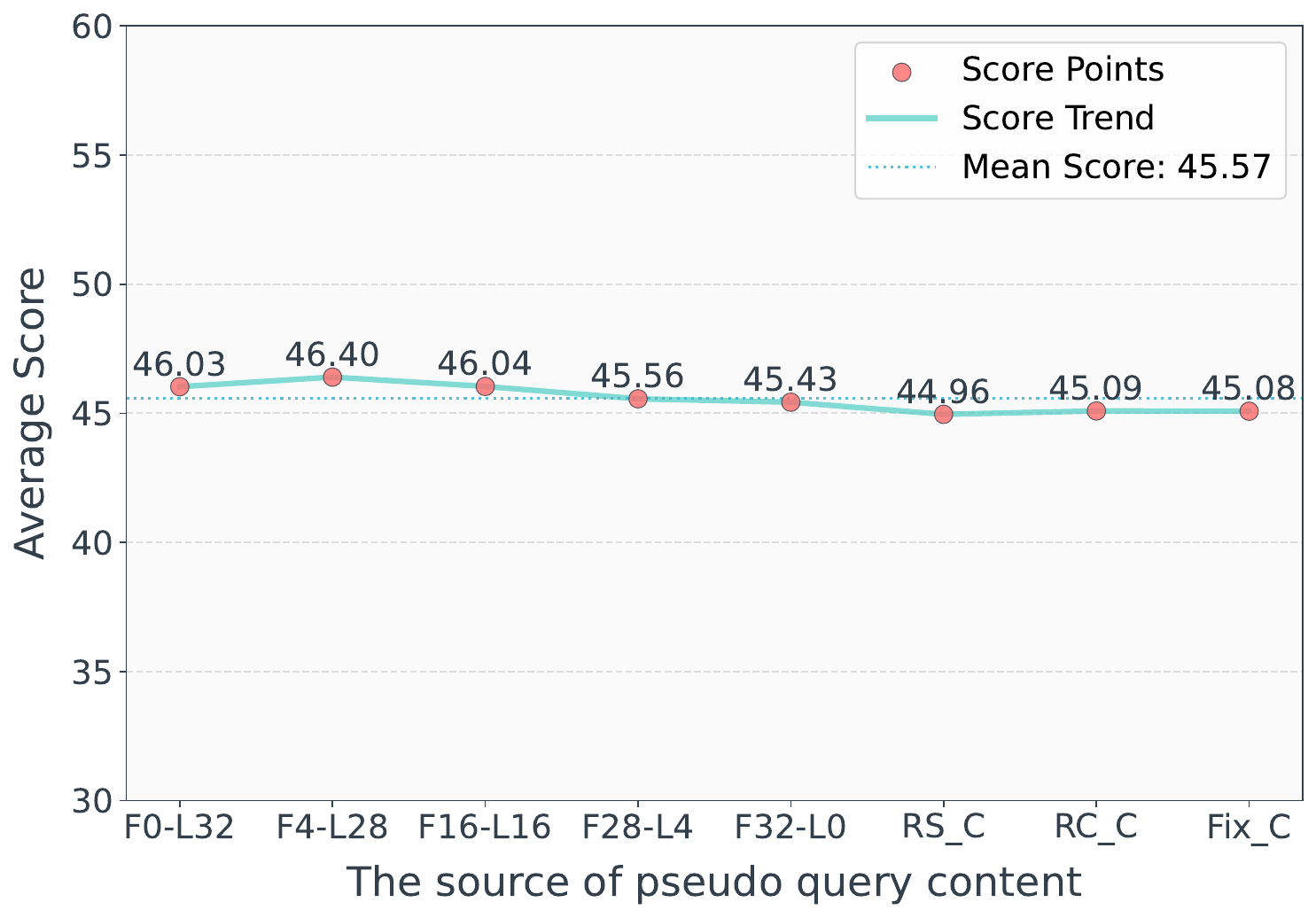}}  
        \vspace{-18pt}
        \caption{The Impact of \texorpdfstring{\(Q_{\text{pseudo}}\)}{Q\_pseudo} Quality}  
        \label{fig:query_quality}
    \end{subfigure}
    \hfill  
    \begin{subfigure}[t]{0.32\textwidth}
        \centering
        \resizebox{\textwidth}{!}{\includegraphics{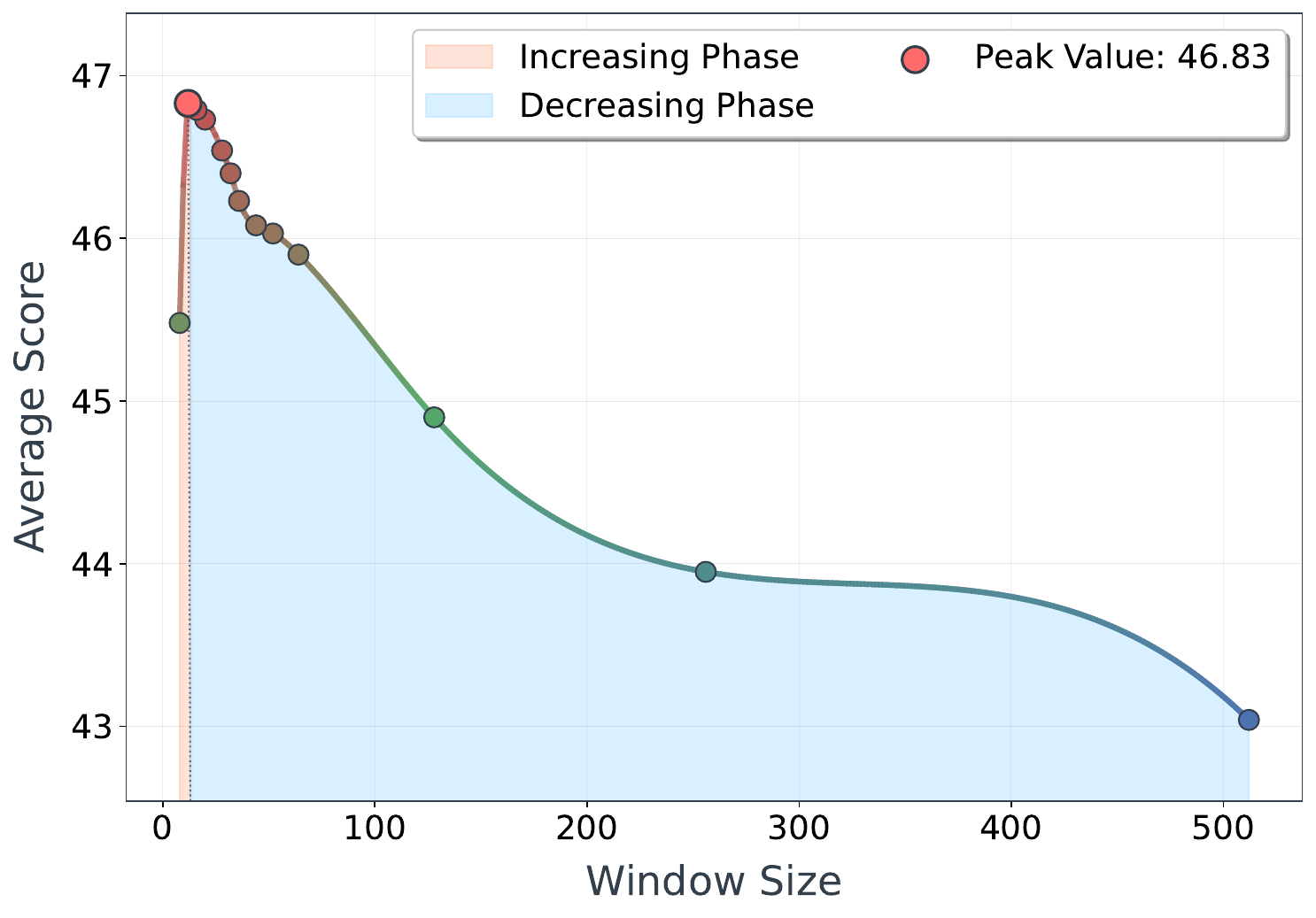}}  
        \vspace{-18pt}
        \caption{The Impact of \texorpdfstring{\(Q_{\text{pseudo}}\)}{Q\_pseudo} Length}  
        \label{fig:query_length}
    \end{subfigure}
    \hfill  
    \begin{subfigure}[t]{0.32\textwidth}
        \centering
        \resizebox{\textwidth}{!}{\includegraphics{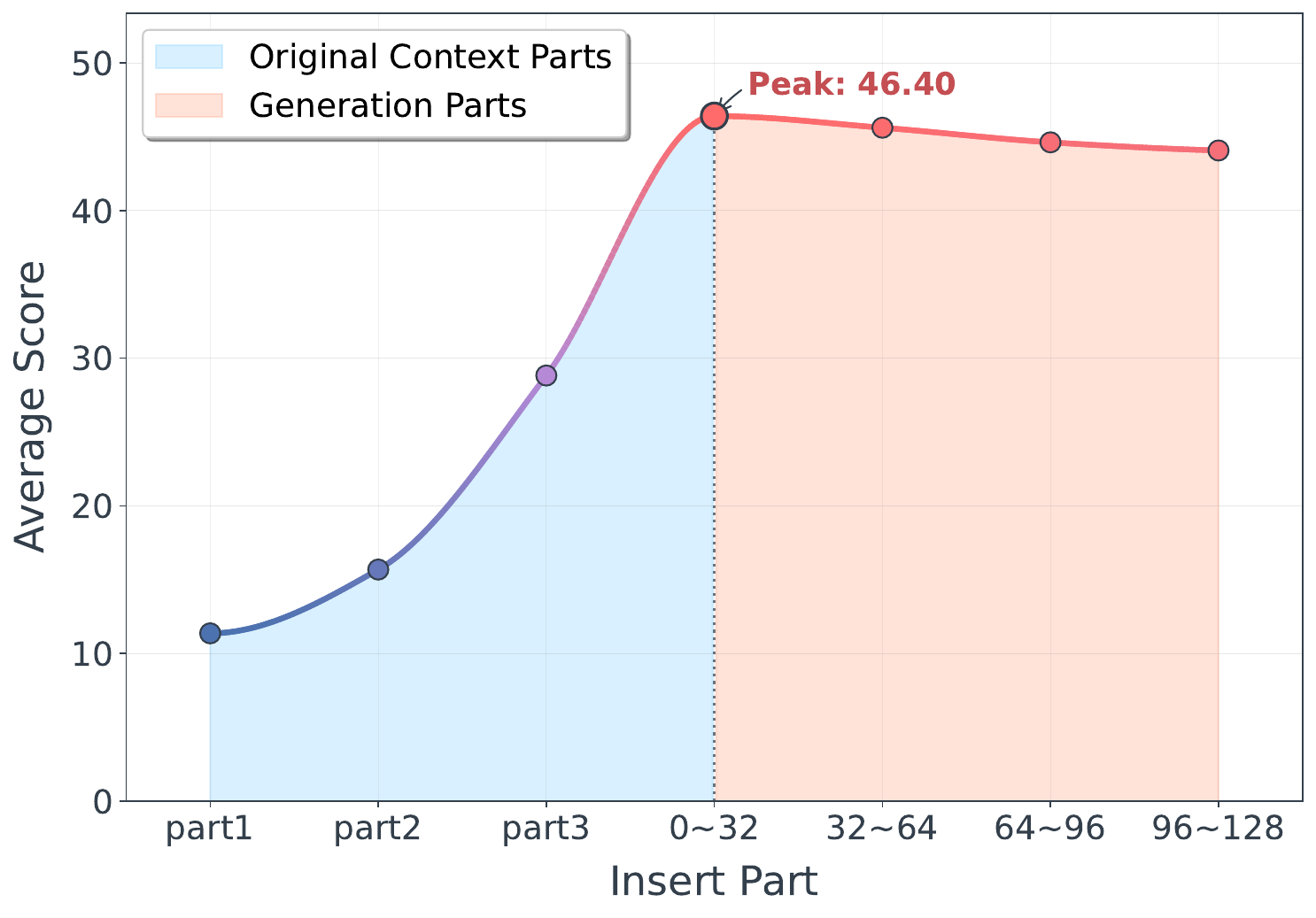}}  
        \vspace{-18pt}
        \caption{The Impact of \texorpdfstring{\(Q_{\text{pseudo}}\)}{Q\_pseudo} Insert Part}  
        \label{fig:query_position}
    \end{subfigure}
    \vspace{-7pt}
\caption{Ablation analysis of pseudo queries with respect to quality, length and insert positions. (a) Performance under a fixed \texorpdfstring{\(Q_{\text{pseudo}}\)}{Q\_pseudo} length $32$, with varying semantic content of \texorpdfstring{\(Q_{\text{pseudo}}\)}{Q\_pseudo}: \texttt{Fm\_Ln} is the concatenation of the first \texttt{m} and the last \texttt{n} tokens from the input context; \texttt{RS\_C} is constructed by concatenating 32 randomly sampled individual tokens from the context; \texttt{RC\_C} is a randomly sampled consecutive span of 32 tokens from the context; and \texttt{Fix\_C} is a fixed, repetitive nonsensical sequence (e.g., \enquote{Sorry, I don't know. Sorry, I don't know…}). (b) Performance under varying observation window sizes $N$. (c) Performance under varying insert positions of \texorpdfstring{\(Q_{\text{pseudo}}\)}{Q\_pseudo}. Left Parts: we divide the original context interval $[0, L_p)$ equally into three segments: \texttt{part1} (beginning), \texttt{part2} (middle), and \texttt{part3} (end). For each segment, we randomly select a continuous sequence of 32 positions and map the position IDs of \texorpdfstring{\(Q_{\text{pseudo}}\)}{Q\_pseudo} to it.
Right Parts: we only move backwards the position IDs of \texorpdfstring{\(Q_{\text{pseudo}}\)}{Q\_pseudo} to continuous 32 positions at different offsets after the context $L_p$ (e.g., $[L_p+0, L_p+32)$), $[L_p+32, L_p+64)$).}
    \vspace{-14pt}
\end{figure*}

\subsection{The Impact of \texorpdfstring{\(Q_{\text{pseudo}}\)}{Q\_pseudo} Semantic Content} 
 To further investigate the practical impact of semantic variation on KV cache compression performance, we conduct an ablation study by evaluating DapQ under a fixed cache budget while altering the semantic content of pseudo queries. As shown in Figure \ref{fig:query_quality}, the average performance remains highly stable (e.g., coefficient of variation $\approx 1\%$), regardless of whether the pseudo-query window is constructed from different semantic contents (e.g., the input's prefix and suffix, random in-context tokens, or nonsensical sequences). This consistency provides compelling empirical evidence that the attention pattern relies significantly on positional information to establish importance distribution, rendering the semantic content a secondary factor.

\subsection{The Impact of \texorpdfstring{\(Q_{\text{pseudo}}\)}{Q\_pseudo} Length}
The length of pseudo queries (the size of an observation window, $N$) is a crucial hyper-parameter, controlling the breadth of the simulated decoding context used for importance estimation. Figure \ref{fig:query_length} reveals a non-monotonic relationship between $N$ and performance, characterized by distinct increasing and decreasing phases.

\textbf{Increasing Phase (Small $N$):} For small window sizes, performance increases sharply as $N$ grows. This is because a small window lacks the contextual breadth to robustly estimate the importance of all relevant tokens, causing high uncertainty in the importance assessment. Adding more pseudo queries can provide a more comprehensive simulation of the decoding process, leading to a more accurate and holistic importance distribution. This expanded window thereby enables the model to identify and retain a greater number of critical tokens for future effective generation.

\textbf{Decreasing Phase (Large $N$):} Beyond a certain point, further increasing $N$ leads to a performance decline. We attribute this to a dilution effect: while initial queries in the window are precisely aligned with the start of decoding, later queries represent increasingly speculative future positions. The attention pattern for these distant positions becomes progressively diffuse and attends to tokens less relevant for initial generation steps, introducing noisier and less reliable signals into the aggregated importance scores. And mutual attention among these queries introduces additional interference, further diverting the focus from critical tokens.

This analysis identifies that the window should be sufficiently sized to capture a representative decoding context but not so large as to dilute the attentional signal. The existence of this optimum further confirms that our method is not relying on a brute-force approach. Instead, it performs a precise and efficient simulation by concentrating on the most relevant segment of the decoding trajectory.

\subsection{The Impact of \texorpdfstring{\(Q_{\text{pseudo}}\)}{Q\_pseudo} Insert Positions}

We observe that deviating from the proposed \texorpdfstring{\(Q_{\text{pseudo}}\)}{Q\_pseudo} placement (i.e., immediately following the prompt) leads to performance degradation. This phenomenon can be explained by two key factors.

\textbf{Insufficient context visibility when inserted within the context:} Under the standard causal attention mask, \texorpdfstring{\(Q_{\text{pseudo}}\)}{Q\_pseudo} placed at position $t$ can only attend to the prefix $[0, t)$ and is unable to access the subsequent context. This conflicts with our goal that \texorpdfstring{\(Q_{\text{pseudo}}\)}{Q\_pseudo} should approximate the attention patterns over the full context as seen by future decoding steps. Consequently, as shown in the left part of Figure~\ref{fig:query_position}, \texorpdfstring{\(Q_{\text{pseudo}}\)}{Q\_pseudo} placed earlier in the context fail to reconstruct global attention patterns and exhibit degraded performance.

\textbf{The assigned position of \texorpdfstring{\(Q_{\text{pseudo}}\)}{Q\_pseudo} shifts backwards away from the correct interval, introduces increasing misalignment in the RoPE embedding space.} Large backward shifts in positions lead to accumulated rotational offsets, causing the \texorpdfstring{\(Q_{\text{pseudo}}\)}{Q\_pseudo} representation space to progressively diverge from those of real queries. This misalignment makes it harder for \texorpdfstring{\(Q_{\text{pseudo}}\)}{Q\_pseudo} to approximate the target attention distribution, explaining the observed performance drop with larger positional shifts, as clearly illustrated in the right part of Figure~\ref{fig:query_position}.
\section{Conclusion}
This work underscores the primacy of positional information over semantic content in constructing query approximations and determining attention patterns. We introduce DapQ, a novel KV cache compression framework that leverages position-aware pseudo queries to simulate the output tokens, thereby establishing an effective observation window for importance assessment. During the prefill stage, it aligns closely with the actual generation context and enables precise token eviction. Extensive experiments demonstrate that DapQ consistently outperforms existing baselines and achieves superior performance in long-context scenarios, particularly under strict memory constraints.



\section*{Limitations}
Although DapQ achieves excellent results, there are still some limitations.

While preliminary experiments indicate that positional information dominates query approximation, semantic content still plays a non-negligible role (as shown in Table \ref{table:query_similarity}: similarity drops to 0.3267 when content is the same but positions differ, whereas it remains at 0.7238 when positions are correct but content is irrelevant). This suggests that further optimizing the semantic content of \texorpdfstring{\(Q_{\text{pseudo}}\)}{Q\_pseudo} to better mirror actual decoding could lead to additional performance gains and enhanced robustness. Future work may explore how to intelligently construct or select semantically more meaningful \texorpdfstring{\(Q_{\text{pseudo}}\)}{Q\_pseudo} content without introducing significant computational overhead, thereby approaching near-lossless performance even under extreme compression scenarios.

The sensitivity of \texorpdfstring{\(Q_{\text{pseudo}}\)}{Q\_pseudo} to positional and semantic information may vary across layers, which could lead to suboptimal compression in certain layers. Future work could explore layer-wise or adaptively-aware \texorpdfstring{\(Q_{\text{pseudo}}\)}{Q\_pseudo} approximation mechanisms to better align with the attention distribution of each layer, thereby further improving overall compression efficiency and generation quality.


\bibliography{custom}

@article{gpt,
  title={Gpt-4 technical report},
  author={Achiam, Josh and Adler, Steven and Agarwal, Sandhini and Ahmad, Lama and Akkaya, Ilge and Aleman, Florencia Leoni and Almeida, Diogo and Altenschmidt, Janko and Altman, Sam and Anadkat, Shyamal and others},
  journal={arXiv preprint arXiv:2303.08774},
  year={2023}
}

@article{clip,
  title={From clip to dino: Visual encoders shout in multi-modal large language models},
  author={Jiang, Dongsheng and Liu, Yuchen and Liu, Songlin and Zhao, Jin'e and Zhang, Hao and Gao, Zhen and Zhang, Xiaopeng and Li, Jin and Xiong, Hongkai},
  journal={arXiv preprint arXiv:2310.08825},
  year={2023}
}

@article{gemini,
  title={Gemini 1.5: Unlocking multimodal understanding across millions of tokens of context},
  author={Team, Gemini and Georgiev, Petko and Lei, Ving Ian and Burnell, Ryan and Bai, Libin and Gulati, Anmol and Tanzer, Garrett and Vincent, Damien and Pan, Zhufeng and Wang, Shibo and others},
  journal={arXiv preprint arXiv:2403.05530},
  year={2024}
}

@article{deepseek,
  title={Deepseek-v3 technical report},
  author={Liu, Aixin and Feng, Bei and Xue, Bing and Wang, Bingxuan and Wu, Bochao and Lu, Chengda and Zhao, Chenggang and Deng, Chengqi and Zhang, Chenyu and Ruan, Chong and others},
  journal={arXiv preprint arXiv:2412.19437},
  year={2024}
}

@inproceedings{sumsurvey,
  title={SumSurvey: An abstractive dataset of scientific survey papers for long document summarization},
  author={Liu, Ran and Liu, Ming and Yu, Min and Zhang, He and Jiang, Jianguo and Li, Gang and Huang, Weiqing},
  booktitle={Findings of the Association for Computational Linguistics ACL 2024},
  pages={9632--9651},
  year={2024}
}

@article{deepseek_coder,
  title={DeepSeek-Coder: When the Large Language Model Meets Programming--The Rise of Code Intelligence},
  author={Guo, Daya and Zhu, Qihao and Yang, Dejian and Xie, Zhenda and Dong, Kai and Zhang, Wentao and Chen, Guanting and Bi, Xiao and Wu, Yu and Li, YK and others},
  journal={arXiv preprint arXiv:2401.14196},
  year={2024}
}

@article{comprehensive,
  title={A comprehensive survey on long context language modeling},
  author={Liu, Jiaheng and Zhu, Dawei and Bai, Zhiqi and He, Yancheng and Liao, Huanxuan and Que, Haoran and Wang, Zekun and Zhang, Chenchen and Zhang, Ge and Zhang, Jiebin and others},
  journal={arXiv preprint arXiv:2503.17407},
  year={2025}
}

@article{llama3,
  title={The llama 3 herd of models},
  author={Grattafiori, Aaron and Dubey, Abhimanyu and Jauhri, Abhinav and Pandey, Abhinav and Kadian, Abhishek and Al-Dahle, Ahmad and Letman, Aiesha and Mathur, Akhil and Schelten, Alan and Vaughan, Alex and others},
  journal={arXiv preprint arXiv:2407.21783},
  year={2024}
}

@article{qwen2,
  title={Qwen2. 5-1m technical report},
  author={Yang, An and Yu, Bowen and Li, Chengyuan and Liu, Dayiheng and Huang, Fei and Huang, Haoyan and Jiang, Jiandong and Tu, Jianhong and Zhang, Jianwei and Zhou, Jingren and others},
  journal={arXiv preprint arXiv:2501.15383},
  year={2025}
}

@article{qwen3,
  title={Qwen3 technical report},
  author={Yang, An and Li, Anfeng and Yang, Baosong and Zhang, Beichen and Hui, Binyuan and Zheng, Bo and Yu, Bowen and Gao, Chang and Huang, Chengen and Lv, Chenxu and others},
  journal={arXiv preprint arXiv:2505.09388},
  year={2025}
}

@article{longbench,
  title={Longbench: A bilingual, multitask benchmark for long context understanding},
  author={Bai, Yushi and Lv, Xin and Zhang, Jiajie and Lyu, Hongchang and Tang, Jiankai and Huang, Zhidian and Du, Zhengxiao and Liu, Xiao and Zeng, Aohan and Hou, Lei and others},
  journal={arXiv preprint arXiv:2308.14508},
  year={2023}
}

@article{longbenchv2,
  title={Longbench v2: Towards deeper understanding and reasoning on realistic long-context multitasks},
  author={Bai, Yushi and Tu, Shangqing and Zhang, Jiajie and Peng, Hao and Wang, Xiaozhi and Lv, Xin and Cao, Shulin and Xu, Jiazheng and Hou, Lei and Dong, Yuxiao and others},
  journal={arXiv preprint arXiv:2412.15204},
  year={2024}
}

@article{needle,
  title={Needle in a haystack-pressure testing llms, 2023},
  author={Kamradt, Gregory},
  journal={URL https://github. com/gkamradt/LLMTest\_NeedleInAHaystack},
  year={2024}
}

@article{ruler,
  title={RULER: What's the Real Context Size of Your Long-Context Language Models?},
  author={Hsieh, Cheng-Ping and Sun, Simeng and Kriman, Samuel and Acharya, Shantanu and Rekesh, Dima and Jia, Fei and Zhang, Yang and Ginsburg, Boris},
  journal={arXiv preprint arXiv:2404.06654},
  year={2024}
}

@article{helmet,
  title={Helmet: How to evaluate long-context language models effectively and thoroughly},
  author={Yen, Howard and Gao, Tianyu and Hou, Minmin and Ding, Ke and Fleischer, Daniel and Izsak, Peter and Wasserblat, Moshe and Chen, Danqi},
  journal={arXiv preprint arXiv:2410.02694},
  year={2024}
}

@article{h2o,
  title={H2o: Heavy-hitter oracle for efficient generative inference of large language models},
  author={Zhang, Zhenyu and Sheng, Ying and Zhou, Tianyi and Chen, Tianlong and Zheng, Lianmin and Cai, Ruisi and Song, Zhao and Tian, Yuandong and R{\'e}, Christopher and Barrett, Clark and others},
  journal={Advances in Neural Information Processing Systems},
  volume={36},
  pages={34661--34710},
  year={2023}
}

@article{streamingllm,
  title={Efficient streaming language models with attention sinks},
  author={Xiao, Guangxuan and Tian, Yuandong and Chen, Beidi and Han, Song and Lewis, Mike},
  journal={arXiv preprint arXiv:2309.17453},
  year={2023}
}

@article{pyramidkv,
  title={Pyramidkv: Dynamic kv cache compression based on pyramidal information funneling},
  author={Cai, Zefan and Zhang, Yichi and Gao, Bofei and Liu, Yuliang and Li, Yucheng and Liu, Tianyu and Lu, Keming and Xiong, Wayne and Dong, Yue and Hu, Junjie and others},
  journal={arXiv preprint arXiv:2406.02069},
  year={2024}
}

@article{lacache,
  title={LaCache: Ladder-Shaped KV Caching for Efficient Long-Context Modeling of Large Language Models},
  author={Shi, Dachuan and Fu, Yonggan and Yuan, Xiangchi and Yu, Zhongzhi and You, Haoran and Li, Sixu and Dong, Xin and Kautz, Jan and Molchanov, Pavlo and others},
  journal={arXiv preprint arXiv:2507.14204},
  year={2025}
}

@article{keepkv,
  title={KeepKV: Eliminating Output Perturbation in KV Cache Compression for Efficient LLMs Inference},
  author={Tian, Yuxuan and Wang, Zihan and Peng, Yebo and Yuan, Aomufei and Wang, Zhiming and Yi, Bairen and Liu, Xin and Cui, Yong and Yang, Tong},
  journal={arXiv preprint arXiv:2504.09936},
  year={2025}
}

@article{lookahead,
  title={Lookahead Q-Cache: Achieving More Consistent KV Cache Eviction via Pseudo Query},
  author={Wang, Yixuan and Ji, Shiyu and Liu, Yijun and Xu, Yuzhuang and Xu, Yang and Zhu, Qingfu and Che, Wanxiang},
  journal={arXiv preprint arXiv:2505.20334},
  year={2025}
}

@article{snapkv,
  title={Snapkv: Llm knows what you are looking for before generation},
  author={Li, Yuhong and Huang, Yingbing and Yang, Bowen and Venkitesh, Bharat and Locatelli, Acyr and Ye, Hanchen and Cai, Tianle and Lewis, Patrick and Chen, Deming},
  journal={Advances in Neural Information Processing Systems},
  volume={37},
  pages={22947--22970},
  year={2024}
}

@article{calidrop,
  title={CaliDrop: KV Cache Compression with Calibration},
  author={Su, Yi and Qiu, Quantong and Zhou, Yuechi and Li, Juntao and Xia, Qingrong and Li, Ping and Duan, Xinyu and Wang, Zhefeng and Zhang, Min},
  journal={arXiv preprint arXiv:2507.19906},
  year={2025}
}

@article{prefix_suffix,
  title={Lost in the middle: How language models use long contexts},
  author={Liu, Nelson F and Lin, Kevin and Hewitt, John and Paranjape, Ashwin and Bevilacqua, Michele and Petroni, Fabio and Liang, Percy},
  journal={arXiv preprint arXiv:2307.03172},
  year={2023}
}

@article{kivi,
  title={Kivi: A tuning-free asymmetric 2bit quantization for kv cache},
  author={Liu, Zirui and Yuan, Jiayi and Jin, Hongye and Zhong, Shaochen and Xu, Zhaozhuo and Braverman, Vladimir and Chen, Beidi and Hu, Xia},
  journal={arXiv preprint arXiv:2402.02750},
  year={2024}
}

@article{kvquant,
  title={Kvquant: Towards 10 million context length llm inference with kv cache quantization},
  author={Hooper, Coleman and Kim, Sehoon and Mohammadzadeh, Hiva and Mahoney, Michael W and Shao, Yakun S and Keutzer, Kurt and Gholami, Amir},
  journal={Advances in Neural Information Processing Systems},
  volume={37},
  pages={1270--1303},
  year={2024}
}

@article{gqa,
  title={Gqa: Training generalized multi-query transformer models from multi-head checkpoints},
  author={Ainslie, Joshua and Lee-Thorp, James and De Jong, Michiel and Zemlyanskiy, Yury and Lebr{\'o}n, Federico and Sanghai, Sumit},
  journal={arXiv preprint arXiv:2305.13245},
  year={2023}
}

@article{kvsharer,
  title={Kvsharer: Efficient inference via layer-wise dissimilar kv cache sharing},
  author={Yang, Yifei and Cao, Zouying and Chen, Qiguang and Qin, Libo and Yang, Dongjie and Zhao, Hai and Chen, Zhi},
  journal={arXiv preprint arXiv:2410.18517},
  year={2024}
}

@article{get,
  title={Get more with less: Synthesizing recurrence with kv cache compression for efficient llm inference},
  author={Dong, Harry and Yang, Xinyu and Zhang, Zhenyu and Wang, Zhangyang and Chi, Yuejie and Chen, Beidi},
  journal={arXiv preprint arXiv:2402.09398},
  year={2024}
}

@article{loki,
  title={Loki: Low-rank keys for efficient sparse attention},
  author={Singhania, Prajwal and Singh, Siddharth and He, Shwai and Feizi, Soheil and Bhatele, Abhinav},
  journal={Advances in Neural Information Processing Systems},
  volume={37},
  pages={16692--16723},
  year={2024}
}

@article{Interpolation,
  title={Extending context window of large language models via positional interpolation},
  author={Chen, Shouyuan and Wong, Sherman and Chen, Liangjian and Tian, Yuandong},
  journal={arXiv preprint arXiv:2306.15595},
  year={2023}
}

@misc{ntk,
  title={Ntk-aware scaled rope allows llama models to have extended (8k+) context size without any fine-tuning and minimal perplexity degradation},
  author={Peng, Bowen and Quesnelle, Jeffrey},
  year={2023}
}

@article{longlora,
  title={Longlora: Efficient fine-tuning of long-context large language models},
  author={Chen, Yukang and Qian, Shengju and Tang, Haotian and Lai, Xin and Liu, Zhijian and Han, Song and Jia, Jiaya},
  journal={arXiv preprint arXiv:2309.12307},
  year={2023}
}

@article{longformer,
  title={Longformer: The long-document transformer},
  author={Beltagy, Iz and Peters, Matthew E and Cohan, Arman},
  journal={arXiv preprint arXiv:2004.05150},
  year={2020}
}

@article{reformer,
  title={Reformer: The efficient transformer},
  author={Kitaev, Nikita and Kaiser, {\L}ukasz and Levskaya, Anselm},
  journal={arXiv preprint arXiv:2001.04451},
  year={2020}
}

@article{linformer,
  title={Linformer: Self-attention with linear complexity},
  author={Wang, Sinong and Li, Belinda Z and Khabsa, Madian and Fang, Han and Ma, Hao},
  journal={arXiv preprint arXiv:2006.04768},
  year={2020}
}

@article{SSMs,
  title={Efficiently modeling long sequences with structured state spaces},
  author={Gu, Albert and Goel, Karan and R{\'e}, Christopher},
  journal={arXiv preprint arXiv:2111.00396},
  year={2021}
}

@article{longmamba,
  title={LongMamba: Enhancing Mamba's Long Context Capabilities via Training-Free Receptive Field Enlargement},
  author={Ye, Zhifan and Xia, Kejing and Fu, Yonggan and Dong, Xin and Hong, Jihoon and Yuan, Xiangchi and Diao, Shizhe and Kautz, Jan and Molchanov, Pavlo and Lin, Yingyan Celine},
  journal={arXiv preprint arXiv:2504.16053},
  year={2025}
}

@article{hcattention,
  title={HCAttention: Extreme KV Cache Compression via Heterogeneous Attention Computing for LLMs},
  author={Yang, Dongquan and Yang, Yifan and Yu, Xiaotian and Qi, Xianbiao and Xiao, Rong},
  journal={arXiv preprint arXiv:2507.19823},
  year={2025}
}

@inproceedings{deepspeed,
  title={Deepspeed-inference: enabling efficient inference of transformer models at unprecedented scale},
  author={Aminabadi, Reza Yazdani and Rajbhandari, Samyam and Awan, Ammar Ahmad and Li, Cheng and Li, Du and Zheng, Elton and Ruwase, Olatunji and Smith, Shaden and Zhang, Minjia and Rasley, Jeff and others},
  booktitle={SC22: International Conference for High Performance Computing, Networking, Storage and Analysis},
  pages={1--15},
  year={2022},
  organization={IEEE}
}

@article{gemodel,
  title={Model tells you what to discard: Adaptive kv cache compression for llms},
  author={Ge, Suyu and Zhang, Yunan and Liu, Liyuan and Zhang, Minjia and Han, Jiawei and Gao, Jianfeng},
  journal={arXiv preprint arXiv:2310.01801},
  year={2023}
}

@article{sepllm,
  title={Sepllm: Accelerate large language models by compressing one segment into one separator},
  author={Chen, Guoxuan and Shi, Han and Li, Jiawei and Gao, Yihang and Ren, Xiaozhe and Chen, Yimeng and Jiang, Xin and Li, Zhenguo and Liu, Weiyang and Huang, Chao},
  journal={arXiv preprint arXiv:2412.12094},
  year={2024}
}

@article{simple,
  title={A Simple and Effective $ L\_2 $ Norm-Based Strategy for KV Cache Compression},
  author={Devoto, Alessio and Zhao, Yu and Scardapane, Simone and Minervini, Pasquale},
  journal={arXiv preprint arXiv:2406.11430},
  year={2024}
}

@article{deepseekv3,
  title={Deepseek-v3 technical report},
  author={Liu, Aixin and Feng, Bei and Xue, Bing and Wang, Bingxuan and Wu, Bochao and Lu, Chengda and Zhao, Chenggang and Deng, Chengqi and Zhang, Chenyu and Ruan, Chong and others},
  journal={arXiv preprint arXiv:2412.19437},
  year={2024}
}

@article{relayattention,
  title={Relayattention for efficient large language model serving with long system prompts},
  author={Zhu, Lei and Wang, Xinjiang and Zhang, Wayne and Lau, Rynson WH},
  journal={arXiv preprint arXiv:2402.14808},
  year={2024}
}

@article{gear,
  title={Gear: An efficient kv cache compression recipe for near-lossless generative inference of llm},
  author={Kang, Hao and Zhang, Qingru and Kundu, Souvik and Jeong, Geonhwa and Liu, Zaoxing and Krishna, Tushar and Zhao, Tuo},
  journal={arXiv preprint arXiv:2403.05527},
  year={2024}
}

@article{palu,
  title={Palu: Compressing kv-cache with low-rank projection},
  author={Chang, Chi-Chih and Lin, Wei-Cheng and Lin, Chien-Yu and Chen, Chong-Yan and Hu, Yu-Fang and Wang, Pei-Shuo and Huang, Ning-Chi and Ceze, Luis and Abdelfattah, Mohamed S and Wu, Kai-Chiang},
  journal={arXiv preprint arXiv:2407.21118},
  year={2024}
}

@article{mqa,
  title={Fast transformer decoding: One write-head is all you need},
  author={Shazeer, Noam},
  journal={arXiv preprint arXiv:1911.02150},
  year={2019}
}

@article{homogeneous,
  title={Homogeneous Keys, Heterogeneous Values: Exploiting Local KV Cache Asymmetry for Long-Context LLMs},
  author={Cui, Wanyun and Xu, Mingwei},
  journal={arXiv preprint arXiv:2506.05410},
  year={2025}
}

@article{sun2024you,
  title={You only cache once: Decoder-decoder architectures for language models},
  author={Sun, Yutao and Dong, Li and Zhu, Yi and Huang, Shaohan and Wang, Wenhui and Ma, Shuming and Zhang, Quanlu and Wang, Jianyong and Wei, Furu},
  journal={Advances in Neural Information Processing Systems},
  volume={37},
  pages={7339--7361},
  year={2024}
}

@article{wu2024layer,
  title={Layer-condensed kv cache for efficient inference of large language models},
  author={Wu, Haoyi and Tu, Kewei},
  journal={arXiv preprint arXiv:2405.10637},
  year={2024}
}

@article{brandon2024reducing,
  title={Reducing transformer key-value cache size with cross-layer attention},
  author={Brandon, William and Mishra, Mayank and Nrusimha, Aniruddha and Panda, Rameswar and Ragan-Kelley, Jonathan},
  journal={Advances in Neural Information Processing Systems},
  volume={37},
  pages={86927--86957},
  year={2024}
}

@article{hydragen,
  title={Hydragen: High-throughput llm inference with shared prefixes},
  author={Juravsky, Jordan and Brown, Bradley and Ehrlich, Ryan and Fu, Daniel Y and R{\'e}, Christopher and Mirhoseini, Azalia},
  journal={arXiv preprint arXiv:2402.05099},
  year={2024}
}

\clearpage
\appendix
\section{Preliminary Experiment Details}
\label{Preliminary Experiment Details}
\subsection{Query similarity comparison under different content and position conditions.}
We conduct the query similarity analysis using a representative example from the GovReport dataset, which is designed for document summarization tasks. The input sequence consists of 4424 tokens. The ground-truth decoding queries are obtained by extracting the first 32 output tokens generated by LLaMA-3-8B-Instruct for this input. The semantic content of these output tokens is: \textcolor{gray}{\enquote{The report discusses the Federal Aviation Administration's (FAA) state block grant pilot program, which is part of its Airport Improvement Program (AIP). The program}}, and they are assigned the positional indices 4424-4455. We then construct a \texorpdfstring{\(Q_{\text{pseudo}}\)}{Q\_pseudo} of length 32 with the repetitive content: \textcolor{gray}{\enquote{Sorry, I don't know. Sorry, I don't know. Sorry, I don't know. Sorry, I don't know. Sorry, I}}. Notably, the positional IDs for the \texorpdfstring{\(Q_{\text{pseudo}}\)}{Q\_pseudo} can be flexibly configured to emulate various decoding scenarios.

\subsubsection{Experimental Details of Query Similarity Comparison}
In Table \ref{table:query_similarity}, we evaluate \texorpdfstring{\(Q_{\text{pseudo}}\)}{Q\_pseudo} by varying two key attributes relative to the ground-truth decoding queries: semantic content and positional assignment. The content is either \textbf{consistent with} the true output (i.e., the actual beginning of the model's summary, \textcolor{gray}{\enquote{The report discusses...}}) or \textbf{different from} it (e.g., a fixed and nonsensical sequence, \textcolor{gray}{\enquote{Sorry, I don't know...}}). Similarly, the positional indices are either \textbf{aligned with} the true future decoding positions (i.e., 4424-4455) or \textbf{deviated from} them (e.g., assigned to a random consecutive span 0-31). The cosine similarity between each set of \texorpdfstring{\(Q_{\text{pseudo}}\)}{Q\_pseudo} and the ground-truth queries is reported under two measurement conditions: \textit{Post ROPE}, which captures the final query representation after the application of Rotary Position Embedding (ROPE), and \textit{Pre ROPE}, which reflects the similarity before the positional encoding is applied.

\subsubsection{Experimental Details of Query Similarity Distribution Analysis}
To quantitatively assess the impact of positional and content variations on query representation, we conduct a large-scale statistical analysis as depicted in Figure \ref{fig:query_similarity_boxplot}. Each box is aggregated from 100 independent trials, providing a stable estimate of the similarity distribution. The \textbf{Different Content (DC)} condition is implemented by randomly sampling 32 tokens from the model's full vocabulary, effectively removing any meaningful semantic correlation with the true output. The \textbf{Different Position (DP)} condition is implemented by assigning a consecutive span of 32 positions randomly sampled from two distinct ranges to introduce positional deviation: a general deviation range (0-4000), which represents a random mismatch within the context window, and an extreme deviation range (0-100), which is specifically chosen to maximize the absolute offset from the correct positions  (i.e., 4424-4455), thereby rigorously testing the hypothesis that positional accuracy is dominant.

\subsection{Experimental Details of Positional Offset Sensitivity}
To systematically quantify the sensitivity of query representations to positional miscalibration, we conduct the analysis presented in Figure \ref{fig:query_similarity_curve}. The experiment investigates how the similarity between \texorpdfstring{\(Q_{\text{pseudo}}\)}{Q\_pseudo} and the true decoding decays based on the absolute offset between their assigned positional indices and the correct future positions. For this purpose, we select input examples of varying context lengths (2k, 4k, 6k, and 8k tokens) from the GovReport dataset. For each context length, we construct \texorpdfstring{\(Q_{\text{pseudo}}\)}{Q\_pseudo} with fixed semantic content (aligned with the true output) but systematically vary their assigned starting position. The x-axis represents this starting position assigned to \texorpdfstring{\(Q_{\text{pseudo}}\)}{Q\_pseudo} (e.g., an x-axis value of 3500 indicates that the 32 \texorpdfstring{\(Q_{\text{pseudo}}\)}{Q\_pseudo} are assigned the consecutive position IDs from 3500 to 3531). The y-axis measures the resulting cosine similarity between the \texorpdfstring{\(Q_{\text{pseudo}}\)}{Q\_pseudo} and the ground-truth decoding queries. This approach allows us to observe the monotonic decay in similarity with increasing positional offset.

\subsection{Generalization Analysis of Positional Dominance}
To further validate the universality of the "Positional Dominance" phenomenon and address potential concerns regarding model-specific or dataset-specific biases, we extended our experimental analysis beyond the Llama-3-8B model and the GovReport dataset. We conducted supplementary experiments using the Qwen2.5-7B-Instruct model across two distinct task domains: Multi-Document QA (using the 2WikiMQA dataset) and Code Completion (using the LCC dataset). Strictly adhering to the experimental configuration outlined in Table \ref{table:query_similarity}, we randomly sampled 100 examples from each dataset to compute the average cosine similarities.  
As shown in Table \ref{table:query_similarity_qwen_lcc} and Table \ref{table:query_similarity_qwen_wiki}, the quantitative results and experimental phenomena for the LCC and 2WikiMQA dataset are consistent with our observations on Llama. This evidence strongly supports the generalization of our central insight.
\begin{table}[t]
\centering
\resizebox{\columnwidth}{!}{%
\begin{tabular}{ccccc}
\toprule
\textbf{Experiment} & \textbf{Content Sim.} & \textbf{Positional Sim.} & \textbf{Post ROPE} & \textbf{Pre ROPE} \\
\midrule
\multirow{1}{*}{SC \& SP} 
& Same
& \multirow{1}{*}{Same} 
& \multirow{1}{*}{\textcolor{green}{1.0000}} & \multirow{1}{*}{\textcolor{green}{1.0000}} \\
\addlinespace
\multirow{1}{*}{DC \& SP} 
& Different
& \multirow{1}{*}{Same} 
& \multirow{1}{*}{\textcolor{green}{0.7142}} & \multirow{1}{*}{\textcolor{green}{0.7142}} \\
\addlinespace
\multirow{1}{*}{SC \& DP} 
& Same
& \multirow{1}{*}{Different} 
& \multirow{1}{*}{\textcolor{red}{0.4270}} & \multirow{1}{*}{\textcolor{green}{0.7341}} \\
\addlinespace
\multirow{1}{*}{DC \& DP} 
& Different
& \multirow{1}{*}{Different} 
& \multirow{1}{*}{\textcolor{red}{0.4113}} & \multirow{1}{*}{\textcolor{green}{0.7224}} \\
\bottomrule
\end{tabular}%
}
\vspace{-8pt}
\caption{Query cosine similarity comparison on the LCC dataset using Qwen2.5-7B-Instruct.}
\label{table:query_similarity_qwen_lcc}
\vspace{-8pt}
\end{table}

\begin{table}[t]
\centering
\resizebox{\columnwidth}{!}{%
\begin{tabular}{ccccc}
\toprule
\textbf{Experiment} & \textbf{Content Sim.} & \textbf{Positional Sim.} & \textbf{Post ROPE} & \textbf{Pre ROPE} \\
\midrule
\multirow{1}{*}{SC \& SP} 
& Same
& \multirow{1}{*}{Same} 
& \multirow{1}{*}{\textcolor{green}{1.0000}} & \multirow{1}{*}{\textcolor{green}{1.0000}} \\
\addlinespace
\multirow{1}{*}{DC \& SP} 
& Different
& \multirow{1}{*}{Same} 
& \multirow{1}{*}{\textcolor{green}{0.7236}} & \multirow{1}{*}{\textcolor{green}{0.7236}} \\
\addlinespace
\multirow{1}{*}{SC \& DP} 
& Same
& \multirow{1}{*}{Different} 
& \multirow{1}{*}{\textcolor{red}{0.4671}} & \multirow{1}{*}{\textcolor{green}{0.7359}} \\
\addlinespace
\multirow{1}{*}{DC \& DP} 
& Different
& \multirow{1}{*}{Different} 
& \multirow{1}{*}{\textcolor{red}{0.4428}} & \multirow{1}{*}{\textcolor{green}{0.7192}} \\
\bottomrule
\end{tabular}%
}
\vspace{-8pt}
\caption{Query cosine similarity comparison on the 2WikiMQA dataset using Qwen2.5-7B-Instruct.}
\label{table:query_similarity_qwen_wiki}
\vspace{-12pt}
\end{table}

\begin{table}[t]
    \centering
    \small 
    \setlength{\tabcolsep}{3pt}
    \renewcommand{\arraystretch}{0.85}
    \resizebox{\columnwidth}{!}{
        \begin{tabular}{lcccccc}
            \toprule
            \multirow{2}{*}{\textbf{Method}} & \multicolumn{6}{c}{\textbf{Batch Size}} \\
            \cmidrule(lr){2-7}
            & 1 & 10 & 20 & 30 & 40 & 50 \\
            \midrule
            FullKV     & 16.87 & 33.74 & 52.49 & 71.24 & OOM   & OOM   \\
            LaCache    & 16.87 & 33.74 & 52.49 & 71.24 & OOM   & OOM   \\
            SLM        & 16.01 & 25.17 & 35.36 & 45.55 & 55.73 & 65.92 \\
            H2O        & 16.01 & 25.17 & 35.36 & 45.55 & 55.73 & 65.92 \\
            PyramidKV  & 16.01 & 25.23 & 35.47 & 45.72 & 55.96 & 66.20 \\
            SnapKV     & 16.01 & 25.17 & 35.36 & 45.55 & 55.73 & 65.92 \\
            DapQ       & 16.01 & 25.21 & 35.43 & 45.65 & 55.87 & 66.02 \\
            \bottomrule
        \end{tabular}
    }
    \vspace{-10pt}
    \caption{Comparison of memory usage (GB) with different batch sizes.}
    \vspace{-12pt}
    \label{tab:memory_comparison}
\end{table}

\section{Complete Main Experiment Results and Details}
\label{Complete Experiment Results}

\subsection{Results and Details on LongBench}
\label{Complete Experiment Results on LongBench}
We comprehensively evaluate the performance of DapQ and baselines on LongBench benchmark shown in Table \ref{table:longbench_all}, with the following setup:
\begin{itemize}[leftmargin=*, nosep] 
    \item \textbf{Models:} LLaMA-3-8B-Instruct, Qwen2.5-7B-Instruct, Qwen3-8B(Reasoning OFF);
    \item \textbf{KV Cache Budgets:} 256, 128, 64 tokens.
\end{itemize}

\subsection{Results and Details on LongBenchV2}
We comprehensively evaluate the performance of DapQ and baselines on LongBenchV2 benchmark shown in Table \ref{table:longbenchv2}, with the following setup:
\begin{itemize}[leftmargin=*, nosep] 
    \item \textbf{Models:} LLaMA-3-8B-Instruct, LLaMA-3.1-8B-Instruct, Qwen2.5-7B-Instruct, Qwen3-8B(Reasoning OFF);
    \item \textbf{KV Cache Budgets:} 128, 64 tokens.
\end{itemize}

\subsection{Results and Details on Ruler}
We comprehensively evaluate the performance of DapQ and baselines on Ruler benchmark shown in Table \ref{table:ruler_llama3}, Table \ref{table:ruler_qwen25}, Table \ref{table:ruler_qwen3}, with the following setup:
\begin{itemize}[leftmargin=*, nosep] 
    \item \textbf{Models:} LLaMA-3-8B-Instruct, Qwen2.5-7B-Instruct, Qwen3-8B(Reasoning OFF);
    \item \textbf{KV Cache Budgets:} 4096, 2048, 1024, 512, 256, 128, 64 tokens.
\end{itemize}

\subsection{Results and Details on HELMET}
We comprehensively evaluate the performance of DapQ and baselines on HELMET benchmark shown in Table \ref{table:helmet}, with the following setup:
\begin{itemize}[leftmargin=*, nosep] 
    \item \textbf{Models:} LLaMA-3-8B-Instruct, Qwen2.5-7B-Instruct;
    \item \textbf{KV Cache Budgets:} 2048, 1024, 512, 256, 128 tokens.
\end{itemize}

\subsection{Results and Details on NIAH}
We comprehensively evaluate the performance of DapQ and baselines on Needle-in-a-Haystack(NIAH) benchmark shown in Table \ref{table:needle}, with the following setup:
\begin{itemize}[leftmargin=*, nosep] 
    \item \textbf{Models:} LLaMA-3-8B-Instruct, LLaMA-3.1-8B-Instruct, Qwen2.5-7B-Instruct, Qwen3-8B(Reasoning OFF);
    \item \textbf{KV Cache Budgets:} 256, 128, 64 tokens.
\end{itemize}

\subsection{Efficiency Evaluation Setup}
We conduct a comprehensive efficiency evaluation of DapQ, focusing on memory usage (Table \ref{tab:memory_comparison}), throughput (Table \ref{tab:throughput_comparison}), and Time-to-First-Token (Table \ref{tab:ttft_comparison}). The specific experimental environment and configurations are detailed as follows:

\begin{itemize}[leftmargin=*, nosep]
    \item \textbf{Hardware \& Environment:} All efficiency experiments are conducted on a single H20 96G GPU. The environment consists of Python 3.10, PyTorch 2.6.0, and Transformers 4.53.0.
    \item \textbf{Attention Kernel:} Utilize Flash-Attention (version 2.7.4) to accelerate attention computation.
    \item \textbf{Inference Framework:} Adopt the native Hugging Face transformers implementation rather than vLLM.
\end{itemize}

\section{Theoretical and Empirical Evidence for Query-Attention Alignment}

\subsection{Theoretical Analysis}

To rigorously verify that position-Based queries enable more accurate token eviction, we derive a formal theorem to establish that the similarity of query vectors directly constrains both the upper bound of the attention score error and the lower bound of the attention distribution similarity. This provides a principled justification that high query similarity necessarily leads to high attention distribution.

\paragraph{(1) Theorem:}
Given a fixed KV set $\{K, V\}$ with $K \in \mathbb{R}^{n \times d_k}$ and $V \in \mathbb{R}^{n \times d_v}$, let two unit query vectors $q, q' \in \mathbb{R}^{1 \times d_k}$ have cosine similarity defined as $\text{sim}(q, q') = q \cdot q'$.
The difference between their corresponding attention scores satisfies:

\begin{equation}
\resizebox{\columnwidth}{!}{$
\lVert \text{Attention}(q, K) - \text{Attention}(q', K) \rVert_1
\le
\frac{
K_{\text{max}} \cdot \sqrt{2(1 - \text{sim}(q, q'))}
}{
\sqrt{d_k}
}
\tag{1}
$}
\end{equation}

\begin{equation}
\resizebox{\columnwidth}{!}{$
\text{sim}(\text{Attention}(q, K), \text{Attention}(q', K)) \ge 1 - \frac{
K_{\text{max}} \cdot \sqrt{2(1 - \text{sim}(q, q'))}
}{
2 \sqrt{d_k}
}
\tag{2}
$}
\end{equation}
where:
\begin{itemize}
    \item $d_k$ is the key-vector dimension;
    \item $K_{\text{max}} = \max_j \|k_j\|$, with $k_j$ the $j$-th key vector;
    \item $\text{Attention}(q, K) = \text{softmax}\left(\frac{qK^T}{\sqrt{d_k}}\right) $
\end{itemize}
and these quantities (i.e., $d_k$ and $K_{\text{max}}$) are constants under the fixed KV set.

\paragraph{(2) Core conclusion:}
\textbf{Inequality (1)} directly establishes a positive correlation between the \textit{cosine similarity of queries} and the \textit{similarity of their attention scores}. Specifically, as $\text{sim}(q, q') \to 1$, the upper bound on the scores difference approaches $0$, meaning that the attention scores produced by $q$ and $q'$ become almost identical. \textbf{Inequality (2)} further quantifies the lower bound constraint that query similarity imposes on attention score similarity — the higher the query similarity, the higher the lower bound on attention score similarity.

\paragraph{(3) Detailed Derivation:} \mbox{} \\

\textbf{Step 1: Define key variables and similarity measures}

\begin{itemize}
    \item \textbf{Cosine similarity of queries.} In modern LLMs, query vectors typically originate from normalization layers (e.g., LayerNorm), ensuring their norms remain a stable constant (i.e., $\|q\| \approx \|q'\| \approx C$). Without loss of generality, we assume unitary magnitude ($C=1$) to simplify the notation. Consequently, their cosine similarity reduces to:
    $$
    \text{sim}(q,q')=\frac{q\cdot q'}{\|q\|\cdot \|q'\|}=q\cdot q'.
    $$
    
    \item \textbf{Pre-softmax scores.} The raw matching score between the query and each key is
    $$
    \small
    s_j=\frac{q\cdot k_j}{\sqrt{d_k}},\quad 
    s'_j=\frac{q'\cdot k_j}{\sqrt{d_k}},\quad (j=1,2,\ldots).
    $$
    
    \item \textbf{Attention weights.} After softmax normalization, the attention weights are
    $$
    \alpha=\text{softmax}(s),\qquad \alpha'=\text{softmax}(s'),
    $$
    where $\alpha_j\ge 0$ and $\sum_j \alpha_j=1$.
\end{itemize}

\textbf{Step 2: Upper bound the difference of pre-softmax scores}

Using the Cauchy–Schwarz inequality $|x\cdot y|\le \|x\| \cdot \|y\|$,
the score difference induced by two queries satisfies
\begin{equation}
\begin{aligned}
|s_j - s'_j|
&= \left|\frac{q\cdot k_j}{\sqrt{d_k}} - \frac{q'\cdot k_j}{\sqrt{d_k}}\right| \\
&= \frac{1}{\sqrt{d_k}}|(q-q')\cdot k_j| \\
&\le \frac{1}{\sqrt{d_k}}\|q-q'\| \cdot \|k_j\|.
\end{aligned}
\tag{3}
\end{equation}

Next, for unit vectors $q$ and $q'$ we have the standard identity
$$
\small
\|q-q'\|^2
= \|q\|^2 + \|q'\|^2 - 2q\cdot q'
= 2\bigl(1-\text{sim}(q,q')\bigr),
$$
hence
$$
\|q-q'\| = \sqrt{2\bigl(1-\text{sim}(q,q')\bigr)}.
$$

Substituting this into the score bound Inequality (3) and taking the maximum key norm
$K_{\text{max}}=\max_j \|k_j\|,$
we obtain
\begin{equation}
\max_j \left| s_j - s'_j \right|
\le
\frac{ K_{\text{max}} \cdot \sqrt{2\bigl(1-\text{sim}(q,q')\bigr)}}{\sqrt{d_k}}
\tag{4}
\end{equation}

\textbf{Step 3: Upper bound the difference of attention weights}

The softmax function is Lipschitz continuous: for any score vectors $s, s'$, we have  
$$
\lVert \text{softmax}(s) - \text{softmax}(s') \rVert_1
\le
\lVert s - s' \rVert_\infty .
$$

We note that this bound corresponds to a global worst-case Lipschitz constant of the softmax function; while tighter, input-dependent bounds can be derived via the Jacobian, the above inequality suffices for establishing a conservative and general theoretical guarantee.

Here, the infinity norm is defined as  
$\lVert s - s' \rVert_\infty = \max_j |s_j - s'_j|$,  
and the $L_1$ norm of the attention-weight difference is  
$\lVert \alpha - \alpha' \rVert_1 = \sum_j |\alpha_j - \alpha'_j|$.

Combining this with Inequality (4), we obtain  
\begin{equation}
\lVert \alpha - \alpha' \rVert_1
\le
\frac{K_{\max} \cdot \sqrt{2\bigl(1-\mathrm{sim}(q,q')\bigr)}}{\sqrt{d_k}}
\tag{5}
\end{equation}

\textbf{Step 4: Lower bound the difference of attention weights}

We adopt a standard similarity measure for attention scores based on the Total Variation Distance:  
$$
\text{sim}(\alpha, \alpha') = 1 - \frac{1}{2} \| \alpha - \alpha' \|_1
$$  
where $\| \alpha - \alpha' \|_1$ takes values in $[0, 2]$. 

First, we scale the L1 difference $[0, 2]$ to $[0, 1]$: $\frac{1}{2} \|\alpha - \alpha'\|_1$. 

Then, by subtracting this scaled value from 1, we convert the "difference" into "similarity":  
\begin{itemize}
    \item Smaller difference $\to$ smaller scaled value $\to$ similarity closer to 1;
    \item Larger difference $\to$ larger scaled value $\to$ similarity closer to 0.
\end{itemize}

The above definition is a standard measure of similarity in the field of probability distributions. Substituting Inequality (5) into this definition yields the final lower bound:  
\begin{equation}
\text{sim}(\alpha, \alpha') \ge 1 - \frac{
K_{\text{max}} \cdot \sqrt{2(1 - \text{sim}(q, q'))}
}{
2 \sqrt{d_k}
} \tag{6}
\end{equation}

\subsection{Empirical Evaluation}

We evaluate the cosine similarity, between Snapkv, DapQ’s observation window attention weights for the input context and ground-truth decoding query’s attention weights for the input context under different window sizes. As shown in the Table \ref{table:query-attention}, across all window sizes, DapQ consistently achieves substantially higher attention-weight similarity compared to SnapKV. This directly demonstrates that better query approximation indeed translates into more accurate attention distributions.

\begin{table*}[h]
\centering
\caption{Cosine similarity of softmax-normalized attention weights for DapQ vs. SnapKV.}
\label{tab:attention_comparison}
    \begin{tabular}{lcccccccc}
    \toprule
    Window Size & 128 & 64 & 32 & 16 & 8 & 4 & 2 & 1 \\
    \midrule
    DapQ        & 0.9785 & 0.9747 & 0.9726 & 0.9713 & 0.9705 & 0.9644 & 0.9615 & 0.8768 \\
    SnapKV      & 0.9463 & 0.9429 & 0.9508 & 0.9466 & 0.9416 & 0.9385 & 0.9332 & 0.8426 \\
    \bottomrule
    \end{tabular}%
\label{table:query-attention}
\end{table*}

\newpage


\begin{table*}[t]
\caption{Performance comparison of different methods across various LLMs on LongBench.}
\centering
\small
\resizebox{\textwidth}{!}{%
\begin{tabular}{>{\bfseries}c >{\raggedright}p{1.2cm} *{14}{c}}
\toprule
&  & \multicolumn{2}{c}{\textbf{Single-Document QA}} & \multicolumn{2}{c}{\textbf{Multi-Document QA}} & \multicolumn{2}{c}{\textbf{Summarization}} & \multicolumn{3}{c}{\textbf{Few-shot Learning}} & \multicolumn{2}{c}{\textbf{Synthetic}} & \multicolumn{2}{c}{\textbf{Code}} &  \\
\cmidrule(lr){3-4}\cmidrule(lr){5-6}\cmidrule(lr){7-8}\cmidrule(lr){9-11}\cmidrule(lr){12-13}\cmidrule(lr){14-15}
& \textbf{Methods} & \rotatebox[origin=c]{30}{\textbf{Qasper}} & \rotatebox[origin=c]{30}{\textbf{MF-en}} & \rotatebox[origin=c]{30}{\textbf{HotpotQA}} & \rotatebox[origin=c]{30}{\textbf{2WikiMQA}} & \rotatebox[origin=c]{30}{\textbf{GovReport}} & \rotatebox[origin=c]{30}{\textbf{MultiNews}} & \rotatebox[origin=c]{30}{\textbf{TREC}} & \rotatebox[origin=c]{30}{\textbf{TriviaQA}} & \rotatebox[origin=c]{30}{\textbf{SAMSum}} & \rotatebox[origin=c]{30}{\textbf{PCount}} & \rotatebox[origin=c]{30}{\textbf{PRe}} & \rotatebox[origin=c]{30}{\textbf{Lcc}} & \rotatebox[origin=c]{30}{\textbf{RB-P}} & \textbf{Avg.} \\
\midrule
& FullKV & 37.72 & 40.64 & 50.17 & 34.88 & 31.03 & 25.64 & 70.00 & 89.85 & 40.55 & 13.30 & 83.67 & 58.96 & 52.71 & 48.39 \\
& \multicolumn{15}{c}{\cellcolor{gray!20}\textbf{KV Cache Size = 256}} \\
\multirow{13}{*}{\rotatebox{90}{Llama3-8B-Instruct}} & H2O & 28.11 & 36.63 & 48.62 & 31.50 & 21.87 & 21.44 & 45.67 & 89.49 & 38.28 & 12.11 & 83.67 & 61.49 & 53.36 & 44.02 \\
& PyramidKV & 30.88 & 38.11 & 50.20 & 33.88 & \textbf{22.54} & 21.84 & 60.00 & 89.26 & 37.07 & \textbf{12.78} & 83.67 & 61.34 & 52.51 & 45.70 \\
& SnapKV & 30.84 & \textbf{38.39} & 49.75 & 33.80 & 22.18 & 21.53 & 57.00 & 89.65 & 36.97 & 12.11 & \textbf{84.00} & 61.78 & 54.92 & 45.61 \\
& DapQ & \textbf{32.55} & 38.18 & \textbf{50.67} & \textbf{34.35} & 22.25 & \textbf{21.89} & \textbf{60.67} & \textbf{90.48} & \textbf{38.34} & 11.78 & 83.67 & \textbf{62.78} & \textbf{55.64} & \textbf{46.40} \\
& \multicolumn{15}{c}{\cellcolor{gray!20}\textbf{KV Cache Size = 128}} \\
& H2O & 25.95 & 36.25 & 48.65 & 31.90 & \textbf{20.79} & 20.30 & 40.00 & 87.29 & 36.25 & 12.33 & \textbf{83.67} & 59.81 & 53.14 & 42.79 \\
& PyramidKV & 28.80 & \textbf{38.29} & 49.52 & 31.60 & 20.67 & 20.55 & 49.00 & 87.68 & 36.73 & \textbf{12.44} & 82.00 & 60.36 & 52.03 & 43.82 \\
& SnapKV & \textbf{29.52} & 37.80 & 49.36 & 32.40 & 19.87 & 20.08 & 47.67 & 87.82 & 35.63 & 11.44 & 82.33 & 61.49 & 52.40 & 43.68 \\
& DapQ & 28.76 & 37.24 & \textbf{50.04} & \textbf{33.59} & 20.47 & \textbf{20.63} & \textbf{50.00} & \textbf{90.06} & \textbf{36.87} & 12.11 & 81.67 & \textbf{61.81} & \textbf{53.92} & \textbf{44.40} \\
& \multicolumn{15}{c}{\cellcolor{gray!20}\textbf{KV Cache Size = 64}} \\
& H2O & 24.02 & 30.83 & 48.27 & 31.70 & \textbf{19.37} & \textbf{19.14} & 37.33 & 86.27 & 35.18 & 7.72 & \textbf{82.33} & 59.20 & \textbf{51.10} & 40.96 \\
& PyramidKV & 22.04 & 31.80 & 47.01 & 31.54 & 15.70 & 16.34 & 39.00 & 76.80 & 32.31 & 10.33 & 79.67 & 55.19 & 47.90 & 38.90 \\
& SnapKV & 25.06 & 32.92 & 47.16 & 31.71 & 16.85 & 17.09 & \textbf{40.67} & 86.02 & 33.99 & 11.78 & 78.00 & 57.95 & 50.91 & 40.78 \\
& DapQ & \textbf{25.99} & \textbf{37.36} & \textbf{49.11} & \textbf{32.88} & 18.46 & 18.70 & 38.67 & \textbf{87.38} & \textbf{35.30} & \textbf{11.89} & 77.67 & \textbf{60.19} & 49.90 & \textbf{41.81} \\
\midrule
& FullKV & 36.50 & 49.70 & 55.91 & 44.70 & 31.64 & 22.84 & 66.33 & 89.34 & 42.49 & 11.00 & 86.33 & 61.97 & 59.97 & 50.67 \\
& \multicolumn{15}{c}{\cellcolor{gray!20}\textbf{KV Cache Size = 256}} \\
\multirow{12}{*}{\rotatebox{90}{Qwen2.5-7B-Instruct}} & H2O & 27.76 & 42.94 & 47.89 & 40.97 & 22.13 & 18.31 & 44.33 & 84.51 & 39.77 & 10.67 & 86.00 & 57.07 & 54.09 & 44.31 \\
& PyramidKV & 29.65 & 46.37 & 49.89 & 40.77 & 20.42 & 16.80 & 53.00 & 87.89 & 39.61 & 10.67 & 86.00 & 52.61 & 49.49 & 44.82 \\
& SnapKV & \textbf{31.12} & \textbf{47.87} & 51.76 & 40.78 & 22.25 & \textbf{18.41} & 53.33 & 87.66 & 39.34 & 10.67 & 86.00 & 56.87 & \textbf{54.36} & 46.19 \\
& DapQ & 30.21 & 45.00 & \textbf{51.92} & \textbf{41.46} & \textbf{22.35} & 18.40 & \textbf{56.67} & \textbf{88.64} & \textbf{39.92} & \textbf{11.00} & \textbf{86.00} & \textbf{58.40} & 53.82 & \textbf{46.45} \\
& \multicolumn{15}{c}{\cellcolor{gray!20}\textbf{KV Cache Size = 128}} \\
& H2O & 26.83 & 37.80 & 45.14 & 39.77 & \textbf{20.13} & 16.64 & 40.67 & 81.43 & 38.56 & 10.67 & \textbf{85.67} & 53.97 & 51.52 & 42.22 \\
& PyramidKV & 26.37 & 43.09 & 46.57 & 39.07 & 18.01 & 15.15 & 43.33 & 84.69 & 38.40 & 10.67 & 84.67 & 49.45 & 46.90 & 42.03 \\
& SnapKV & \textbf{27.46} & 41.81 & 48.62 & 41.40 & 19.42 & 16.19 & 42.33 & 83.91 & \textbf{38.89} & 10.67 & 85.00 & 52.14 & 51.39 & 43.02 \\
& DapQ & 26.81 & \textbf{43.12} & \textbf{49.62} & \textbf{41.36} & 19.89 & \textbf{16.68} & \textbf{47.00} & \textbf{84.92} & 38.07 & \textbf{11.00} & 85.00 & \textbf{54.46} & \textbf{51.70} & \textbf{43.82} \\
& \multicolumn{15}{c}{\cellcolor{gray!20}\textbf{KV Cache Size = 64}} \\
& H2O & 24.33 & 32.36 & 44.94 & 39.06 & \textbf{17.96} & \textbf{15.05} & 37.33 & 82.76 & 35.27 & 10.67 & \textbf{85.00} & \textbf{49.44} & 45.52 & 39.98 \\
& PyramidKV & 22.36 & 35.19 & 43.51 & 38.08 & 14.04 & 11.10 & 37.33 & 84.81 & 35.58 & 10.67 & 80.33 & 44.67 & 42.22 & 38.45 \\
& SnapKV & 22.90 & 40.66 & 45.56 & 40.28 & 15.42 & 12.03 & 37.67 & \textbf{85.11} & 36.92 & 10.67 & 82.00 & 46.16 & 45.75 & 40.09 \\
& DapQ & \textbf{25.58} & \textbf{42.65} & \textbf{49.66} & \textbf{41.25} & 16.90 & 14.05 & \textbf{39.67} & 84.16 & \textbf{35.49} & \textbf{11.00} & 80.67 & 49.38 & \textbf{46.77} & \textbf{41.33} \\
\midrule
& FullKV & 36.87 & 53.67 & 57.67 & 44.73 & 33.39 & 23.69 & 71.67 & 91.79 & 42.07 & 11.98 & 86.67 & 70.64 & 59.26 & 52.60 \\
& \multicolumn{15}{c}{\cellcolor{gray!20}\textbf{KV Cache Size = 256}} \\
\multirow{12}{*}{\rotatebox{90}{Qwen3-8B}} & H2O & 27.80 & 44.92 & 51.37 & 40.50 & 22.38 & 18.26 & 46.33 & 90.04 & 38.98 & 12.33 & 86.67 & 66.11 & 53.93 & 46.12 \\
& PyramidKV & 30.41 & 47.81 & 51.00 & 41.37 & 22.36 & 17.41 & 61.67 & 90.96 & 37.65 & 13.00 & 86.33 & 67.01 & 50.11 & 47.47 \\
& SnapKV & \textbf{32.40} & 49.29 & 54.38 & 41.40 & 23.79 & 18.99 & \textbf{63.00} & 91.07 & 38.57 & \textbf{14.67} & 86.67 & \textbf{67.99} & 53.77 & 48.92 \\
& DapQ & 32.14 & \textbf{50.78} & \textbf{54.79} & \textbf{44.47} & \textbf{24.16} & \textbf{19.01} & 62.67 & \textbf{91.15} & \textbf{39.61} & 14.17 & \textbf{86.67} & 67.20 & \textbf{53.83} & \textbf{49.28} \\
& \multicolumn{15}{c}{\cellcolor{gray!20}\textbf{KV Cache Size = 128}} \\
& H2O & 26.60 & 41.37 & 48.10 & 39.85 & 20.83 & 17.27 & 41.33 & 90.21 & 38.16 & \textbf{11.72} & \textbf{86.67} & 65.22 & 52.77 & 44.22 \\
& PyramidKV & 26.22 & 40.48 & 48.41 & 39.46 & 18.99 & 15.10 & 48.33 & 89.31 & 36.92 & 9.67 & 86.67 & 60.82 & 48.78 & 43.78 \\
& SnapKV & \textbf{29.41} & 46.43 & 51.20 & 41.66 & 20.37 & 16.64 & 51.33 & \textbf{91.07} & 37.37 & 11.00 & 86.67 & 65.36 & 51.65 & 46.17 \\
& DapQ & 29.10 & \textbf{47.15} & \textbf{53.84} & \textbf{43.00} & \textbf{21.11} & \textbf{17.27} & \textbf{54.00} & 90.45 & \textbf{38.50} & 11.67 & 86.00 & \textbf{66.29} & \textbf{52.03} & \textbf{46.95} \\
& \multicolumn{15}{c}{\cellcolor{gray!20}\textbf{KV Cache Size = 64}} \\
& H2O & 25.55 & 38.94 & 46.66 & 39.27 & \textbf{18.55} & \textbf{15.23} & 39.00 & 88.13 & 35.98 & 9.67 & \textbf{86.67} & 59.48 & \textbf{48.95} & 42.47 \\
& PyramidKV & 25.32 & 40.44 & 46.61 & 39.20 & 16.25 & 12.93 & \textbf{44.67} & 88.27 & 34.63 & 11.33 & 83.67 & 59.73 & 46.48 & 42.27 \\
& SnapKV & 25.09 & 39.89 & 46.58 & 39.38 & 15.28 & 12.38 & 42.67 & 87.93 & 35.12 & 11.33 & 84.33 & 57.96 & 46.49 & 41.88 \\
& DapQ & \textbf{25.78} & \textbf{43.17} & \textbf{49.84} & \textbf{41.28} & 17.16 & 13.95 & 43.00 & \textbf{88.97} & \textbf{36.00} & \textbf{12.67} & 83.00 & \textbf{60.31} & 46.90 & \textbf{43.23} \\
\bottomrule
\end{tabular}
}
\label{table:longbench_all}
\end{table*}




\begin{table*}[t]
\caption{Performance comparison of different methods across various LLMs on LongBenchv2. For DapQ, the pseudo queries are constructed via prefix-suffix concatenation: using the first 8 and last 24 tokens for LLaMA series models, and the first 2 and last 30 tokens for Qwen models. Notably, several compressed methods surpass the FullKV baseline. We attribute this phenomenon to the noise reduction mechanism of cache eviction. By selectively retaining critical tokens, these methods effectively reduce noise and sparsify the context, potentially leading to more focused and effective model reasoning. This effect is particularly pronounced in long-context benchmarks.}
\centering
\small
\scriptsize 
\begin{tabular}{>{\bfseries}c l *{6}{c}} 
\toprule
& & \multicolumn{2}{c}{\textbf{Difficulty}} & \multicolumn{3}{c}{\textbf{Length}} & \\
\cmidrule(lr){3-4} \cmidrule(lr){5-7}
\multirow{1}{*}{\textbf{LLMs}} & \multirow{1}{*}{\textbf{Methods}} & \textbf{Easy} & \textbf{Hard} & \textbf{Short} & \textbf{Medium} & \textbf{Long} & \multirow{1}{*}{\textbf{Overall}} \\ 
\midrule
\multirow{11}{*}{\makecell[c]{Llama3-8B\\Instruct}} & FullKV & 28.65 & 26.37 & 32.22 & 24.19 & 25.00 & 27.24 \\
& \multicolumn{7}{c}{\cellcolor{gray!20}\textbf{KV Cache Size = 128}} \\
& H2O & \textbf{32.29} & 25.40 & 32.22 & 26.98 & 23.15 & 28.03 \\
& PyramidKV & 30.21 & 24.44 & 31.67 & 26.05 & 19.44 & 26.64 \\
& SnapKV & 30.73 & 25.72 & 32.22 & 26.05 & 23.15 & 27.63 \\
& DapQ & 30.73 & \textbf{27.65} & \textbf{33.33} & \textbf{27.91} & \textbf{23.15} & \textbf{28.83} \\
& \multicolumn{7}{c}{\cellcolor{gray!20}\textbf{KV Cache Size = 64}} \\
& H2O & 30.73 & 24.76 & 28.89 & 26.51 & 25.00 & 27.04 \\
& PyramidKV & 27.08 & 23.47 & 24.44 & 27.44 & 20.37 & 24.85 \\
& SnapKV & \textbf{31.25} & 22.51 & 23.89 & 26.98 & 26.85 & 25.84 \\
& DapQ & 30.73 & \textbf{29.26} & \textbf{31.11} & \textbf{28.84} & \textbf{29.63} & \textbf{29.82} \\
\midrule
\multirow{11}{*}{\makecell[c]{Llama3.1-8B\\Instruct}} & FullKV & 25.00 & 28.62 & 31.67 & 25.58 & 23.15 & 27.24 \\
& \multicolumn{7}{c}{\cellcolor{gray!20}\textbf{KV Cache Size = 128}} \\
& H2O & 26.56 & 29.58 & 34.44 & 25.58 & 24.07 & 28.43 \\
& PyramidKV & \textbf{29.17} & 28.94 & 32.22 & 27.44 & 26.85 & 29.03 \\
& SnapKV & 27.08 & 29.90 & 33.89 & 26.05 & 25.93 & 28.83 \\
& DapQ & 27.08 & \textbf{30.55} & \textbf{34.44} & \textbf{27.44} & \textbf{24.07} & \textbf{29.22} \\
& \multicolumn{7}{c}{\cellcolor{gray!20}\textbf{KV Cache Size = 64}} \\
& H2O & 23.44 & 27.01 & 30.56 & 23.72 & 21.30 & 25.65 \\
& PyramidKV & 28.12 & 28.62 & 33.89 & 24.19 & \textbf{27.78} & 28.43 \\
& SnapKV & 25.00 & 26.69 & 29.44 & 24.65 & 23.15 & 26.04 \\
& DapQ & \textbf{29.17} & \textbf{28.94} & \textbf{33.89} & \textbf{26.98} & 25.00 & \textbf{29.03} \\
\midrule
\multirow{11}{*}{\makecell[c]{Qwen2.5-7B\\Instruct}} & FullKV & 28.65 & 27.33 & 30.56 & 27.44 & 24.07 & 27.83 \\
& \multicolumn{7}{c}{\cellcolor{gray!20}\textbf{KV Cache Size = 128}} \\
& H2O & 28.65 & \textbf{27.65} & 30.56 & 27.91 & \textbf{24.07} & 28.03 \\
& PyramidKV & 29.17 & 25.40 & 29.44 & 26.51 & 23.15 & 26.84 \\
& SnapKV & \textbf{29.69} & 26.37 & 30.56 & 27.91 & 22.22 & 27.63 \\
& DapQ & 29.17 & 27.33 & \textbf{30.56} & \textbf{28.37} & 23.15 & \textbf{28.03} \\
& \multicolumn{7}{c}{\cellcolor{gray!20}\textbf{KV Cache Size = 64}} \\
& H2O & 28.65 & 26.37 & 31.11 & 26.51 & 22.22 & 27.24 \\
& PyramidKV & \textbf{31.25} & 27.97 & 32.22 & 28.37 & 25.93 & 29.22 \\
& SnapKV & 30.73 & 27.33 & \textbf{32.78} & 27.44 & 24.07 & 28.63 \\
& DapQ & 30.73 & \textbf{28.30} & 31.67 & \textbf{28.37} & \textbf{26.85} & \textbf{29.22} \\
\midrule
\multirow{11}{*}{\makecell[c]{Qwen3-8B}} & FullKV & 31.25 & 28.30 & 33.33 & 25.58 & 30.56 & 29.42 \\
& \multicolumn{7}{c}{\cellcolor{gray!20}\textbf{KV Cache Size = 128}} \\
& H2O & 34.38 & 27.97 & 35.56 & 25.58 & \textbf{31.48} & 30.42 \\
& PyramidKV & 34.38 & 27.33 & \textbf{36.11} & 26.05 & 27.78 & 30.02 \\
& SnapKV & \textbf{34.90} & 27.33 & 34.44 & 26.05 & 31.48 & 30.22 \\
& DapQ & 33.85 & \textbf{28.30} & 33.89 & \textbf{27.44} & 30.56 & \textbf{30.42} \\
& \multicolumn{7}{c}{\cellcolor{gray!20}\textbf{KV Cache Size = 64}} \\
& H2O & 35.42 & 27.65 & 34.44 & 28.84 & 27.78 & 30.62 \\
& PyramidKV & \textbf{38.02} & 26.69 & 34.44 & 28.37 & 30.56 & 31.01 \\
& SnapKV & 36.46 & 27.33 & 33.33 & \textbf{29.30} & 29.63 & 30.82 \\
& DapQ & 35.94 & \textbf{28.62} & \textbf{36.67} & 26.51 & \textbf{32.41} & \textbf{31.41} \\
\bottomrule
\end{tabular}
\label{table:longbenchv2}
\end{table*}




\begin{table*}[t]
\centering
\caption{Performance comparison of different methods across various kv cache size on Ruler for llama3-8B-Instruct.}
\small
\resizebox{\textwidth}{!}{%
\begin{tabular}{>{\bfseries}c >{\raggedright}p{2.2cm} *{12}{c}}
\toprule
& & \multicolumn{3}{c}{\textbf{Single NIAH}} & \multicolumn{3}{c}{\textbf{Multi-key NIAH}} & & & & & & \\
\cmidrule(lr){3-5}\cmidrule(lr){6-8}
\textbf{LLM} & \textbf{Methods} & \rotatebox[origin=c]{30}{S-NIAH-1} & \rotatebox[origin=c]{30}{S-NIAH-2} & \rotatebox[origin=c]{30}{S-NIAH-3} & \rotatebox[origin=c]{30}{MK-NIAH-1} & \rotatebox[origin=c]{30}{MK-NIAH-2} & \rotatebox[origin=c]{30}{MK-NIAH-3} & \rotatebox[origin=c]{30}{MQ-NIAH} & \rotatebox[origin=c]{30}{MV-NIAH} & \rotatebox[origin=c]{30}{CWE} & \rotatebox[origin=c]{30}{FWE} & \rotatebox[origin=c]{30}{VT} & \textbf{AVG} \\
\midrule
\multirow{43}{*}{\rotatebox{90}{Llama3-8B-Instruct}} & FullKV & 100 & 100 & 100 & 99.2 & 91.8 & 95.8 & 99.75 & 97.5 & 97.82 & 82.93 & 98.32 & 96.65 \\
& \multicolumn{13}{c}{\cellcolor{gray!20}\textbf{KV Cache Size = 2048}} \\
& LaCache & 21 & 27.4 & 1.8 & 29.6 & 29 & 3.2 & 12.2 & 6.45 & 86.08 & 86.07 & 11.04 & 28.53 \\
& SLM & 26.6 & 25.4 & 25.4 & 22.6 & 24.2 & 17.8 & 24.1 & 24.3 & 4.9 & 78.6 & 21.16 & 26.82 \\
& H2O & 100 & 90 & 14 & 78 & 47 & 12.6 & 74.7 & 45 & 90.04 & 79.33 & 97.68 & 66.21 \\
& PyramidKV & 100 & 100 & 40.8 & 99 & 72.2 & 17.4 & 99 & 96.3 & 83.25 & 67.87 & 98.2 & 79.46 \\
& SnapKV & 100 & 98.4 & 61.4 & 99 & 69 & 19.8 & 98.8 & 94.45 & 90.34 & 73.13 & 97.56 & 81.99 \\
& DapQ & 100 & 99 & 96 & 99.2 & 67.2 & 24.4 & 99.05 & 95.45 & 90.68 & 75 & 97.64 & \textbf{85.78} \\
& \multicolumn{13}{c}{\cellcolor{gray!20}\textbf{KV Cache Size = 1024}} \\
& LaCache & 0.2 & 4.8 & 2.4 & 4.2 & 4 & 0 & 2.3 & 2.45 & 55.02 & 86.87 & 1.56 & 14.89 \\
& SLM & 13.4 & 13.8 & 11.4 & 11.4 & 13.2 & 10.8 & 10.95 & 11.3 & 0.38 & 75.07 & 8.16 & 16.35 \\
& H2O & 98.2 & 75.8 & 8 & 62 & 38.8 & 4.8 & 49.05 & 10.6 & 63.4 & 75.07 & 92.2 & 52.54 \\
& PyramidKV & 100 & 98.2 & 6.2 & 98.6 & 47.8 & 2 & 97 & 90.8 & 56.14 & 61.53 & 97.08 & 68.67 \\
& SnapKV & 100 & 96.8 & 15.4 & 98.4 & 44.8 & 3.4 & 96.45 & 87.9 & 65.42 & 64.27 & 96.48 & 69.94 \\
& DapQ & 100 & 98.4 & 85.8 & 99.2 & 41.8 & 6.2 & 97.4 & 92.5 & 65.74 & 69.33 & 96.76 & \textbf{77.56} \\
& \multicolumn{13}{c}{\cellcolor{gray!20}\textbf{KV Cache Size = 512}} \\
& LaCache & 0 & 0.2 & 0 & 2.6 & 0.4 & 0 & 0.05 & 1.4 & 16.52 & 78.8 & 0.08 & 9.10 \\
& SLM & 3.6 & 6.2 & 5.6 & 6.6 & 6.2 & 5.4 & 6.25 & 6.75 & 0.18 & 75.33 & 1.08 & 11.20 \\
& H2O & 88.4 & 63.8 & 2.4 & 45 & 25.4 & 1.4 & 24.4 & 2.85 & 46.56 & 64.6 & 69.64 & 39.50 \\
& PyramidKV & 100 & 95.6 & 0 & 97.4 & 35 & 0.2 & 91.8 & 73.5 & 23.56 & 52.8 & 92.96 & 60.26 \\
& SnapKV & 100 & 95.6 & 1.4 & 96.8 & 30.4 & 0.4 & 91.1 & 71.65 & 30.82 & 53.73 & 94.48 & 60.58 \\
& DapQ & 100 & 97.8 & 59.6 & 98.6 & 29.8 & 1 & 91.95 & 82.9 & 27.16 & 61.27 & 95.2 & \textbf{67.75} \\
& \multicolumn{13}{c}{\cellcolor{gray!20}\textbf{KV Cache Size = 256}} \\
& LaCache & 0 & 0 & 0 & 0 & 0 & 0 & 0 & 0 & 3.64 & 58.2 & 0 & 5.62 \\
& SLM & 1.2 & 1.2 & 1.2 & 2 & 3.4 & 2.4 & 2.3 & 2.45 & 0.14 & 78.6 & 0 & 8.63 \\
& H2O & 67.2 & 56.8 & 2.4 & 25.4 & 15.6 & 0 & 9.05 & 1.25 & 31.1 & 48.2 & 20.64 & 25.24 \\
& PyramidKV & 100 & 94.8 & 0 & 89.6 & 29.4 & 0 & 73.5 & 35.15 & 9.42 & 39.8 & 75.8 & 49.77 \\
& SnapKV & 100 & 95 & 0 & 90.2 & 26 & 0 & 76.5 & 36.3 & 13.66 & 45.2 & 91.56 & 52.22 \\
& DapQ & 100 & 97.6 & 23 & 97.8 & 19.8 & 0 & 79.7 & 55 & 12.8 & 51.87 & 88.04 & \textbf{56.87} \\
& \multicolumn{13}{c}{\cellcolor{gray!20}\textbf{KV Cache Size = 128}} \\
& LaCache & 0 & 0 & 0 & 0 & 0 & 0 & 0 & 0 & 0.58 & 8.4 & 0 & 0.82 \\
& SLM & 0.6 & 1.2 & 1.2 & 2 & 2 & 0 & 2.25 & 2.45 & 0.2 & 44.93 & 0 & 5.17 \\
& H2O & 41.4 & 38.8 & 2.4 & 14.8 & 2.8 & 0 & 2.2 & 0.3 & 18.06 & 13 & 7.88 & 12.88 \\
& PyramidKV & 99.2 & 91 & 0 & 68.2 & 33.2 & 0 & 26.5 & 9.7 & 2.38 & 25.6 & 18.76 & 34.05 \\
& SnapKV & 98.8 & 89 & 0 & 60.2 & 35.4 & 0 & 17.25 & 7.45 & 5.18 & 28.53 & 13.24 & 32.28 \\
& DapQ & 99.6 & 97.6 & 1.4 & 94.4 & 21.4 & 0 & 28.85 & 20.05 & 4.32 & 30.33 & 6.84 & \textbf{36.80} \\
& \multicolumn{13}{c}{\cellcolor{gray!20}\textbf{KV Cache Size = 64}} \\
& LaCache & 0 & 0 & 0 & 0 & 0 & 0 & 0 & 0 & 0.18 & 0.67 & 0 & 0.08 \\
& SLM & 0 & 0 & 0 & 0 & 0 & 0 & 0 & 0 & 0.06 & 27.53 & 0 & 2.51 \\
& H2O & 22 & 21.2 & 0 & 3.8 & 0.2 & 0 & 0.4 & 0.25 & 5.7 & 0.07 & 3.52 & 5.19 \\
& PyramidKV & 48.8 & 47.2 & 0 & 13.4 & 7.2 & 0 & 0.4 & 0.25 & 0.08 & 0 & 0.6 & 10.72 \\
& SnapKV & 58.8 & 65.4 & 0 & 20.4 & 14 & 0 & 0.75 & 0.4 & 0.14 & 0.07 & 2.28 & 14.75 \\
& DapQ & 85.2 & 87.4 & 0 & 26.2 & 19.6 & 0 & 3.65 & 1.35 & 0.78 & 0.33 & 3.6 & \textbf{20.74} \\
\bottomrule
\end{tabular}
}
\label{table:ruler_llama3}
\end{table*}

\clearpage

\begin{table*}[t]
\centering
\caption{Performance comparison of different methods across various kv cache size on Ruler for Qwen2.5-7B-Instruct.}
\small
\resizebox{\textwidth}{!}{%
\begin{tabular}{>{\bfseries}c >{\raggedright}p{2.2cm} *{12}{c}}
\toprule
& & \multicolumn{3}{c}{\textbf{Single NIAH}} & \multicolumn{3}{c}{\textbf{Multi-key NIAH}} & & & & & & \\
\cmidrule(lr){3-5}\cmidrule(lr){6-8}
\textbf{LLM} & \textbf{Methods} & \rotatebox[origin=c]{30}{S-NIAH-1} & \rotatebox[origin=c]{30}{S-NIAH-2} & \rotatebox[origin=c]{30}{S-NIAH-3} & \rotatebox[origin=c]{30}{MK-NIAH-1} & \rotatebox[origin=c]{30}{MK-NIAH-2} & \rotatebox[origin=c]{30}{MK-NIAH-3} & \rotatebox[origin=c]{30}{MQ-NIAH} & \rotatebox[origin=c]{30}{MV-NIAH} & \rotatebox[origin=c]{30}{CWE} & \rotatebox[origin=c]{30}{FWE} & \rotatebox[origin=c]{30}{VT} & \textbf{AVG} \\
\midrule
\multirow{50}{*}{\rotatebox{90}{Qwen2.5-7B-Instruct}} & FullKV & 100 & 99.8 & 99.8 & 99.8 & 98 & 93.2 & 99.8 & 93.9 & 77.38 & 87.67 & 95.36 & 94.97 \\
& \multicolumn{13}{c}{\cellcolor{gray!20}\textbf{KV Cache Size = 4096}} \\
& LaCache & 3 & 1.8 & 2.4 & 4 & 4.8 & 3 & 3.3 & 2.25 & 62.78 & 87.73 & 6.32 & 16.49 \\
& SLM & 26.4 & 28 & 27 & 24.4 & 19.6 & 11.6 & 26.1 & 27.1 & 36.96 & 91.53 & 22.84 & 31.05 \\
& H2O & 100 & 98.4 & 24 & 96.8 & 19.4 & 9.4 & 85.75 & 68.15 & 67.56 & 92.33 & 93.92 & 68.70 \\
& PyramidKV & 100 & 99.4 & 41.8 & 99.4 & 19.8 & 3.6 & 91.6 & 83.4 & 47.66 & 91.07 & 93.96 & 70.15 \\
& SnapKV & 100 & 99.8 & 86 & 99.8 & 39.6 & 13 & 97.15 & 88.6 & 67.72 & 91.2 & 93.64 & 79.68 \\
& DapQ & 100 & 99.2 & 82 & 99 & 56.8 & 23.6 & 97.25 & 82.15 & 68.04 & 92.53 & 94.52 & \textbf{81.37} \\
& \multicolumn{13}{c}{\cellcolor{gray!20}\textbf{KV Cache Size = 2048}} \\
& LaCache & 0.2 & 1.6 & 0 & 2.4 & 0.4 & 0 & 0 & 1.2 & 36.74 & 87.33 & 0.68 & 11.87 \\
& SLM & 13.8 & 13.4 & 11 & 11.4 & 9 & 6.2 & 10.8 & 11.3 & 5.56 & 94.67 & 9.6 & 17.88 \\
& H2O & 98 & 90.8 & 7.8 & 90.2 & 7.2 & 3.8 & 65.5 & 35 & 55.78 & 93.07 & 89.72 & 57.90 \\
& PyramidKV & 99.4 & 97.2 & 9.2 & 97.8 & 11 & 0.8 & 75.9 & 55 & 28.7 & 94.73 & 96.04 & 60.52 \\
& SnapKV & 100 & 99.2 & 47.6 & 98.8 & 25 & 2.4 & 92.25 & 79.85 & 55.72 & 95.33 & 95.6 & 71.98 \\
& DapQ & 100 & 96.4 & 53 & 97.2 & 43.2 & 7 & 93 & 69.4 & 56.44 & 95.6 & 94.44 & \textbf{73.24} \\
& \multicolumn{13}{c}{\cellcolor{gray!20}\textbf{KV Cache Size = 1024}} \\
& LaCache & 0 & 1.6 & 2.4 & 3.2 & 0 & 0 & 1.8 & 2.35 & 13.56 & 84.47 & 0 & 9.94 \\
& SLM & 4.4 & 6.2 & 5.6 & 6.6 & 4 & 4.2 & 6.2 & 6.7 & 0.26 & 96.93 & 3.16 & 13.11 \\
& H2O & 96.4 & 73.8 & 2.4 & 78.2 & 2.8 & 1.6 & 38.2 & 11.15 & 42.12 & 87.53 & 71.16 & 45.94 \\
& PyramidKV & 99 & 91 & 0.6 & 91.6 & 3.6 & 0 & 48.25 & 25.1 & 12.08 & 95.2 & 93.32 & 50.89 \\
& SnapKV & 99.4 & 98.4 & 14.4 & 97.8 & 13.8 & 0.8 & 78.1 & 58.7 & 37.94 & 96.4 & 92.48 & 62.57 \\
& DapQ & 99.8 & 91.8 & 18.4 & 94.2 & 28.4 & 2.4 & 82.15 & 50.8 & 37.34 & 97 & 93.64 & \textbf{63.27} \\
& \multicolumn{13}{c}{\cellcolor{gray!20}\textbf{KV Cache Size = 512}} \\
& LaCache & 0 & 0 & 0 & 0 & 0 & 0 & 0 & 0 & 5 & 71.47 & 0 & 6.95 \\
& SLM & 1.6 & 1.2 & 1.2 & 2 & 2.2 & 2.6 & 2.3 & 2.4 & 0.26 & 98.87 & 0.36 & 10.45 \\
& H2O & 91 & 53.4 & 2.4 & 52.4 & 0.6 & 0.8 & 14 & 3.75 & 25.64 & 64.87 & 39.92 & 31.71 \\
& PyramidKV & 96.8 & 74.8 & 0 & 62.6 & 1 & 0 & 15.6 & 8.6 & 2.52 & 82.47 & 59.84 & 36.75 \\
& SnapKV & 99 & 89.6 & 1 & 93.8 & 5 & 0.2 & 56.8 & 29.55 & 22.1 & 93.33 & 92.12 & 52.95 \\
& DapQ & 99.6 & 82.8 & 3.4 & 87.4 & 16.6 & 0.4 & 62.4 & 29.75 & 21.9 & 95.2 & 87.52 & \textbf{53.36} \\
& \multicolumn{13}{c}{\cellcolor{gray!20}\textbf{KV Cache Size = 256}} \\
& LaCache & 0 & 0 & 0 & 0 & 0 & 0 & 0 & 0 & 1.58 & 21.27 & 0 & 2.08 \\
& SLM & 0.6 & 1.2 & 1.2 & 3.6 & 0.6 & 0 & 2.3 & 2.4 & 0.22 & 98.07 & 0 & 10.02 \\
& H2O & 63.8 & 22.4 & 2.4 & 13.6 & 0.6 & 0 & 4.25 & 1.5 & 12.46 & 31.13 & 8.76 & 14.63 \\
& PyramidKV & 67.4 & 37.6 & 0 & 16.6 & 0.8 & 0 & 0.75 & 0.6 & 0.36 & 60.67 & 23.96 & 18.98 \\
& SnapKV & 97.6 & 73.4 & 0 & 69.6 & 2 & 0 & 18.95 & 5.8 & 10.52 & 81.6 & 54.84 & 37.66 \\
& DapQ & 98.8 & 63.2 & 0.2 & 71.4 & 10.8 & 0 & 21.4 & 6.35 & 10.7 & 83.73 & 56.96 & \textbf{38.5} \\
& \multicolumn{13}{c}{\cellcolor{gray!20}\textbf{KV Cache Size = 128}} \\
& LaCache & 0 & 0 & 0 & 0 & 0 & 0 & 0 & 0 & 0.6 & 2.67 & 0 & 0.30 \\
& SLM & 0.4 & 1.4 & 0 & 2 & 0.2 & 0 & 2.3 & 2.4 & 0.34 & 96.13 & 0 & 9.56 \\
& H2O & 6 & 3.4 & 0 & 3.6 & 0.2 & 0 & 0.05 & 2.05 & 7.26 & 3.4 & 0.48 & 2.40 \\
& PyramidKV & 4.6 & 6 & 0 & 3.2 & 0 & 0 & 0 & 0 & 0.26 & 21.27 & 2.48 & 3.44 \\
& SnapKV & 57.2 & 30.8 & 0 & 16.8 & 0.6 & 0 & 0.4 & 0.3 & 1.02 & 39.27 & 5.64 & 13.82 \\
& DapQ & 58 & 31.4 & 0 & 39.2 & 3.2 & 0 & 0.6 & 0.9 & 1.46 & 34.73 & 9.96 & \textbf{16.31} \\
& \multicolumn{13}{c}{\cellcolor{gray!20}\textbf{KV Cache Size = 64}} \\
& LaCache & 0 & 0 & 0 & 0 & 0 & 0 & 0 & 0 & 0.7 & 0.67 & 0 & 0.12 \\
& SLM & 0 & 0 & 0 & 0 & 0 & 0 & 0 & 0 & 0.38 & 0 & 0.04 & 0.04 \\
& H2O & 0.2 & 0 & 0 & 0 & 0 & 0 & 0 & 0 & 0.98 & 0 & 0.04 & 0.11 \\
& PyramidKV & 0 & 0 & 0 & 0 & 0 & 0 & 0 & 0 & 0.2 & 0 & 0 & 0.02 \\
& SnapKV & 0 & 0 & 0 & 0.4 & 0 & 0 & 0 & 0 & 0.28 & 0.07 & 0.44 & 0.11 \\
& DapQ & 5.4 & 2.6 & 0 & 3.2 & 0.4 & 0 & 0 & 0 & 0.28 & 2.2 & 1.16 & \textbf{1.39} \\
\bottomrule
\end{tabular}
}

\label{table:ruler_qwen25}
\end{table*}

\clearpage

\begin{table*}[t]
\centering
\caption{Performance comparison of different methods across various kv cache size on Ruler for Qwen3-8B.}
\small
\resizebox{\textwidth}{!}{%
\begin{tabular}{>{\bfseries}c >{\raggedright}p{2.2cm} *{12}{c}}
\toprule
& & \multicolumn{3}{c}{\textbf{Single NIAH}} & \multicolumn{3}{c}{\textbf{Multi-key NIAH}} & & & & & & \\
\cmidrule(lr){3-5}\cmidrule(lr){6-8}
\textbf{LLM} & \textbf{Methods} & \rotatebox[origin=c]{30}{S-NIAH-1} & \rotatebox[origin=c]{30}{S-NIAH-2} & \rotatebox[origin=c]{30}{S-NIAH-3} & \rotatebox[origin=c]{30}{MK-NIAH-1} & \rotatebox[origin=c]{30}{MK-NIAH-2} & \rotatebox[origin=c]{30}{MK-NIAH-3} & \rotatebox[origin=c]{30}{MQ-NIAH} & \rotatebox[origin=c]{30}{MV-NIAH} & \rotatebox[origin=c]{30}{CWE} & \rotatebox[origin=c]{30}{FWE} & \rotatebox[origin=c]{30}{VT} & \textbf{AVG} \\
\midrule
\multirow{50}{*}{\rotatebox{90}{Qwen3-8B}} & FullKV & 100 & 100 & 100 & 99.6 & 99.6 & 99.6 & 99.9 & 99.75 & 83.98 & 90.67 & 100 & 97.55 \\
& \multicolumn{13}{c}{\cellcolor{gray!20}\textbf{KV Cache Size = 4096}} \\
& LaCache & 17.6 & 14.6 & 13.2 & 16.8 & 20.2 & 7.4 & 14.75 & 6.2 & 62.46 & 68.47 & 12.48 & 23.11 \\
& SLM & 26.6 & 28 & 27 & 24.4 & 19.6 & 18.4 & 26.15 & 27.1 & 36.62 & 93 & 22.96 & 31.80 \\
& H2O & 100 & 99.8 & 22 & 99.8 & 84.8 & 32.4 & 99.5 & 89.9 & 46.74 & 93 & 99.92 & 78.90 \\
& PyramidKV & 100 & 100 & 24.8 & 99.6 & 91 & 51.2 & 99.9 & 99.55 & 57.4 & 92.87 & 100 & 83.30 \\
& SnapKV & 100 & 100 & 75.2 & 99.8 & 96 & 58 & 99.9 & 99.75 & 72.7 & 93.27 & 100 & 90.42 \\
& DapQ & 100 & 100 & 96.4 & 99.8 & 93 & 58.4 & 99.9 & 99.85 & 71.82 & 93.6 & 100 & \textbf{92.07} \\
& \multicolumn{13}{c}{\cellcolor{gray!20}\textbf{KV Cache Size = 2048}} \\
& LaCache & 2 & 1.6 & 2.4 & 3.4 & 1.2 & 0 & 2.45 & 0.9 & 43.86 & 69.4 & 2.04 & 11.75 \\
& SLM & 13.8 & 13.4 & 11 & 11.4 & 9.4 & 9 & 10.85 & 11.3 & 10.38 & 95.47 & 9.72 & 18.70 \\
& H2O & 100 & 99.4 & 7.8 & 97 & 64.8 & 10.6 & 94.7 & 48.3 & 28.48 & 94.53 & 99.68 & 67.75 \\
& PyramidKV & 100 & 100 & 1.6 & 100 & 79.4 & 19 & 99.9 & 93.6 & 35.46 & 95.6 & 100 & 74.96 \\
& SnapKV & 100 & 100 & 20 & 100 & 91.6 & 30.4 & 99.85 & 97.75 & 49.59 & 96 & 100 & 80.47 \\
& DapQ & 100 & 100 & 55 & 100 & 88.05 & 31 & 99.95 & 95.15 & 45.76 & 96.07 & 100 & \textbf{82.82} \\
& \multicolumn{13}{c}{\cellcolor{gray!20}\textbf{KV Cache Size = 1024}} \\
& LaCache & 0.2 & 1.6 & 2.4 & 3.2 & 0.4 & 0 & 2.45 & 1.05 & 15.86 & 62.07 & 0.32 & 8.14 \\
& SLM & 4.6 & 6.2 & 5.6 & 6.6 & 3.8 & 5 & 6.25 & 6.7 & 0.8 & 96.47 & 3.24 & 13.21 \\
& H2O & 97.8 & 92 & 2.4 & 81.6 & 38.2 & 2.4 & 72 & 15.85 & 23.98 & 96.2 & 93.92 & 56.03 \\
& PyramidKV & 100 & 99.8 & 0 & 98.6 & 61.8 & 2.4 & 97.75 & 70.2 & 17.84 & 92.47 & 99.12 & 67.27 \\
& SnapKV & 100 & 99.6 & 1 & 99.6 & 79.2 & 13 & 99.25 & 83.05 & 26.52 & 96.27 & 98.84 & 72.39 \\
& DapQ & 100 & 99.8 & 14.2 & 99.6 & 75.4 & 14.8 & 99.75 & 78.25 & 23.78 & 97.73 & 99.12 & \textbf{72.95} \\
& \multicolumn{13}{c}{\cellcolor{gray!20}\textbf{KV Cache Size = 512}} \\
& LaCache & 0 & 1.6 & 0 & 3 & 0 & 0 & 0 & 0.8 & 7.06 & 26.07 & 0.08 & 3.51 \\
& SLM & 1.8 & 4 & 1.2 & 3.8 & 2.4 & 3 & 3.75 & 4.4 & 0.76 & 97.6 & 0.56 & 11.21 \\
& H2O & 90.4 & 66.2 & 2.4 & 57.2 & 15 & 0.8 & 30.6 & 5.1 & 14.8 & 89.2 & 65.56 & 39.75 \\
& PyramidKV & 99 & 97.2 & 0 & 84.4 & 40.8 & 0.2 & 70.7 & 30 & 6.2 & 70.27 & 70.92 & 51.79 \\
& SnapKV & 99.6 & 99.2 & 0 & 97.2 & 60.6 & 2 & 94.35 & 47.25 & 14.78 & 93.6 & 98.64 & 64.29 \\
& DapQ & 100 & 99.6 & 5.8 & 95.4 & 68.6 & 5.2 & 94.85 & 42.45 & 12.38 & 96.27 & 96.08 & \textbf{65.15} \\
& \multicolumn{13}{c}{\cellcolor{gray!20}\textbf{KV Cache Size = 256}} \\
& LaCache & 0 & 0 & 0 & 0 & 0 & 0 & 0 & 0 & 1.7 & 7.93 & 0 & 0.88 \\
& SLM & 0.8 & 1.2 & 1.2 & 2 & 2.4 & 0 & 2.3 & 2.4 & 0.72 & 95 & 0 & 9.82 \\
& H2O & 68 & 21.4 & 2.4 & 19 & 2.6 & 0 & 6.15 & 1.4 & 11.2 & 61.27 & 19.2 & 19.33 \\
& PyramidKV & 94.8 & 60.4 & 0 & 41.6 & 14.4 & 0 & 5.8 & 5.6 & 1.22 & 34.07 & 24 & 25.63 \\
& SnapKV & 97.4 & 87.8 & 0 & 81.6 & 35.2 & 0 & 47.2 & 13 & 9.7 & 84.47 & 72.32 & 48.06 \\
& DapQ & 100 & 88.8 & 0 & 75.4 & 59.4 & 0.2 & 47.55 & 10.2 & 6.1 & 88 & 58.04 & \textbf{48.52} \\
& \multicolumn{13}{c}{\cellcolor{gray!20}\textbf{KV Cache Size = 128}} \\
& LaCache & 0 & 0 & 0 & 0 & 0 & 0 & 0 & 0 & 0.46 & 0.2 & 0 & 0.06 \\
& SLM & 0.6 & 1.2 & 1.2 & 2 & 0.2 & 0 & 2.3 & 2.4 & 0.64 & 96 & 0 & 9.69 \\
& H2O & 19.8 & 2.4 & 1.6 & 3.6 & 0.2 & 0 & 0.05 & 0.25 & 8.4 & 22.53 & 3.44 & 5.66 \\
& PyramidKV & 11.4 & 2.8 & 0 & 0 & 1.2 & 0 & 0 & 0 & 0.84 & 4.6 & 3.68 & 2.23 \\
& SnapKV & 68.8 & 31.2 & 0 & 2.6 & 3.2 & 0 & 0.15 & 0.45 & 2.06 & 40.33 & 5.24 & 14.00 \\
& DapQ & 97.8 & 34 & 0 & 7 & 20.8 & 0 & 0.2 & 0.1 & 1.1 & 40.53 & 5.47 & \textbf{18.82} \\
& \multicolumn{13}{c}{\cellcolor{gray!20}\textbf{KV Cache Size = 64}} \\
& LaCache & 0 & 0 & 0 & 0 & 0 & 0 & 0 & 0 & 0.72 & 0 & 0 & 0.07 \\
& SLM & 0 & 0 & 0 & 0 & 0 & 0 & 0 & 0 & 0.4 & 32.33 & 0 & \textbf{2.98} \\
& H2O & 0 & 0 & 0 & 0 & 0 & 0 & 0 & 0 & 3.42 & 0 & 1.24 & 0.42 \\
& PyramidKV & 0 & 0 & 0 & 0.2 & 0 & 0 & 0 & 0 & 0.78 & 0.4 & 0.56 & 0.18 \\
& SnapKV & 0 & 0 & 0 & 0 & 0 & 0 & 0 & 0 & 0.8 & 0.07 & 0.88 & 0.16 \\
& DapQ & 0.8 & 0.2 & 0 & 0 & 1.2 & 0 & 0 & 0 & 1 & 2.4 & 1.68 & 0.66 \\
\bottomrule
\end{tabular}
}
\label{table:ruler_qwen3}
\end{table*}





\begin{table*}[!htbp]
\caption{Performance comparison of different methods across various LLMs on sub-task categories of the HELMET benchmark.}
\centering
\large
\renewcommand{\arraystretch}{1.2} 
\resizebox{\textwidth}{!}{%
\begin{tabular}{l l *{12}{c}}
\toprule
& & \multicolumn{5}{c}{\textbf{ICL}} & \multicolumn{2}{c}{\textbf{LONGQA}} & \multicolumn{4}{c}{\textbf{RAG}} & \\
& & \multicolumn{5}{c}{\textit{exact\_match}} & \textit{f1} & \textit{rougeL\_f1} & \multicolumn{4}{c}{\textit{substring\_exact\_match}} & \\
\cmidrule(lr){3-7} \cmidrule(lr){8-9} \cmidrule(lr){10-13}
& \textbf{Methods} & \rotatebox[origin=c]{30}{\textbf{icl\_banking77}} & \rotatebox[origin=c]{30}{\textbf{icl\_clinic150}} & \rotatebox[origin=c]{30}{\textbf{icl\_nlu}} & \rotatebox[origin=c]{30}{\textbf{icl\_trec\_coarse}} & \rotatebox[origin=c]{30}{\textbf{icl\_trec\_fine}} & \rotatebox[origin=c]{30}{\textbf{narrativeqa}} & \rotatebox[origin=c]{30}{\textbf{infbench\_qa\_eng}} & \rotatebox[origin=c]{30}{\textbf{kilt\_hotpotqa}} & \rotatebox[origin=c]{30}{\textbf{kilt\_nq}} & \rotatebox[origin=c]{30}{\textbf{kilt\_popqa\_3}} & \rotatebox[origin=c]{30}{\textbf{kilt\_triviaqa}} & \textbf{Avg.} \\
\midrule
\multirow{24}{*}{\rotatebox{90}{\textbf{Llama3-8B-Instruct}}} & FullKV & 38.60 & 73.60 & 76.20 & 39.40 & 25.00 & 12.00 & 16.56 & 52.00 & 42.17 & 47.67 & 80.00 & 45.75 \\
\cmidrule{2-14}
& \multicolumn{13}{c}{\cellcolor{gray!20}\textbf{KV Cache Size = 1024}} \\
& H2O & \textbf{32.00} & 64.00 & 70.60 & 36.00 & \textbf{21.20} & 11.30 & 15.28 & 47.67 & 44.17 & 47.67 & 81.50 & 42.85 \\
& PyramidKV & 24.80 & 55.40 & 68.60 & 29.60 & 16.00 & 11.61 & 16.00 & 52.33 & 44.17 & 48.17 & 81.33 & 40.73 \\
& SnapKV & 23.00 & 54.40 & 70.60 & 29.00 & 18.20 & 11.04 & 17.07 & \textbf{52.67} & 43.83 & 48.33 & 81.67 & 40.89 \\
& DapQ & 26.80 & \textbf{64.40} & \textbf{72.60} & \textbf{39.80} & 15.40 & \textbf{11.81} & \textbf{16.67} & 50.00 & \textbf{45.00} & \textbf{48.33} & \textbf{81.83} & \textbf{42.97} \\
\cmidrule{2-14}
& \multicolumn{13}{c}{\cellcolor{gray!20}\textbf{KV Cache Size = 512}} \\
& H2O & 18.00 & 51.80 & 63.00 & 28.80 & 18.00 & 10.65 & 14.45 & 48.33 & 44.17 & 48.00 & 82.00 & 38.84 \\
& PyramidKV & 17.40 & 39.80 & 56.00 & 22.20 & 12.20 & \textbf{11.69} & 15.87 & 50.67 & 44.17 & 48.00 & 82.17 & 36.38 \\
& SnapKV & 17.60 & 41.40 & 63.60 & 22.20 & 14.20 & 11.34 & 16.67 & \textbf{50.67} & 44.50 & 47.17 & 82.83 & 37.47 \\
& DapQ & \textbf{21.60} & \textbf{52.80} & \textbf{64.00} & \textbf{39.20} & \textbf{15.40} & 11.18 & \textbf{16.89} & 49.00 & \textbf{46.00} & \textbf{48.17} & \textbf{82.17} & \textbf{40.58} \\
\cmidrule{2-14}
& \multicolumn{13}{c}{\cellcolor{gray!20}\textbf{KV Cache Size = 256}} \\
& H2O & 11.60 & 35.40 & \textbf{54.20} & 23.40 & 11.20 & 11.21 & 12.82 & 46.67 & 42.33 & 47.00 & 80.67 & 34.23 \\
& PyramidKV & 14.00 & 27.40 & 36.20 & 17.00 & 10.00 & 11.36 & 14.94 & \textbf{52.00} & 42.83 & 48.00 & 81.17 & 32.26 \\
& SnapKV & 16.80 & 27.80 & 47.60 & 17.80 & 9.00 & 11.52 & 14.76 & 50.67 & 42.67 & 46.83 & 81.67 & 33.37 \\
& DapQ & \textbf{17.00} & \textbf{37.40} & 47.00 & \textbf{38.40} & \textbf{13.80} & \textbf{11.66} & \textbf{15.70} & 48.67 & \textbf{44.00} & \textbf{48.17} & \textbf{81.83} & \textbf{36.69} \\
\cmidrule{2-14}
& \multicolumn{13}{c}{\cellcolor{gray!20}\textbf{KV Cache Size = 128}} \\
& H2O & 6.80 & 20.00 & \textbf{32.80} & 18.00 & 6.20 & 10.76 & 12.52 & 45.33 & 41.00 & 46.00 & 81.33 & 29.16 \\
& PyramidKV & 11.20 & 21.80 & 19.60 & 14.80 & 8.20 & 10.70 & 14.03 & 48.00 & 39.33 & 46.50 & 83.50 & 28.88 \\
& SnapKV & 13.80 & 19.60 & 21.40 & 14.60 & 8.80 & 10.92 & 13.66 & 46.00 & 38.83 & 47.07 & 84.33 & 29.00 \\
& DapQ & \textbf{16.40} & \textbf{23.40} & 25.60 & \textbf{31.00} & \textbf{13.20} & \textbf{10.99} & \textbf{15.14} & \textbf{49.33} & \textbf{41.50} & \textbf{47.33} & \textbf{84.67} & \textbf{32.60} \\
\midrule
\multirow{24}{*}{\rotatebox{90}{\textbf{Qwen2.5-7B-Instruct}}} & FullKV & 74.00 & 71.00 & 53.80 & 75.60 & 31.80 & 20.53 & 30.33 & 56.00 & 49.83 & 57.67 & 86.67 & 55.20 \\
\cmidrule{2-14}
& \multicolumn{13}{c}{\cellcolor{gray!20}\textbf{KV Cache Size = 2048}} \\
& H2O & 63.20 & 54.00 & 51.00 & \textbf{79.40} & 31.40 & \textbf{20.37} & 28.74 & 52.00 & \textbf{49.33} & 58.17 & 87.00 & 52.24 \\
& PyramidKV & 68.40 & 61.40 & 53.20 & 77.40 & 33.20 & 19.46 & 28.52 & 53.67 & 48.00 & \textbf{60.83} & 85.33 & 53.58 \\
& SnapKV & 68.40 & 62.00 & 51.40 & 78.00 & 33.20 & 20.35 & 29.41 & 55.67 & 47.83 & 58.33 & 86.67 & 53.75 \\
& DapQ & \textbf{71.00} & \textbf{67.80} & \textbf{55.00} & 76.00 & \textbf{33.60} & 19.05 & \textbf{29.46} & \textbf{56.33} & 48.83 & 58.50 & \textbf{87.00} & \textbf{54.78} \\
\cmidrule{2-14}
& \multicolumn{13}{c}{\cellcolor{gray!20}\textbf{KV Cache Size = 1024}} \\
& H2O & 44.60 & 33.20 & 28.00 & \textbf{79.40} & 25.40 & 19.52 & 27.24 & 50.67 & \textbf{48.50} & 57.00 & 85.50 & 45.37 \\
& PyramidKV & 58.40 & 40.60 & 39.00 & 78.20 & 29.20 & 20.30 & 28.11 & 50.67 & 44.50 & \textbf{61.00} & 84.50 & 48.59 \\
& SnapKV & 56.80 & 42.10 & 38.00 & 75.00 & 30.80 & \textbf{20.53} & 28.05 & 56.00 & 48.00 & 58.33 & 85.37 & 49.00 \\
& DapQ & \textbf{66.00} & \textbf{56.40} & \textbf{49.80} & 76.20 & \textbf{31.20} & 20.20 & \textbf{28.76} & \textbf{54.33} & 46.67 & 57.50 & \textbf{85.67} & \textbf{52.07} \\
\cmidrule{2-14}
& \multicolumn{13}{c}{\cellcolor{gray!20}\textbf{KV Cache Size = 512}} \\
& H2O & 28.60 & 19.40 & 17.00 & \textbf{76.20} & 17.80 & 18.51 & 26.88 & 51.27 & 45.33 & 57.67 & 85.33 & 40.36 \\
& PyramidKV & 45.00 & 20.00 & 22.80 & 70.60 & 25.80 & \textbf{21.91} & 26.60 & 48.67 & 43.50 & \textbf{59.67} & 82.83 & 42.49 \\
& SnapKV & 44.60 & 23.60 & 22.20 & 71.80 & 26.80 & 20.17 & 27.85 & 53.00 & \textbf{47.17} & 58.17 & 85.83 & 43.74 \\
& DapQ & \textbf{52.00} & \textbf{43.20} & \textbf{40.80} & 72.20 & \textbf{28.40} & 20.54 & \textbf{29.50} & \textbf{54.00} & 45.33 & 57.33 & \textbf{85.83} & \textbf{48.10} \\
\cmidrule{2-14}
& \multicolumn{13}{c}{\cellcolor{gray!20}\textbf{KV Cache Size = 256}} \\
& H2O & 21.20 & 12.00 & 9.20 & \textbf{74.40} & 14.40 & 18.26 & 26.12 & 48.67 & 43.00 & 58.67 & 83.33 & 37.20 \\
& PyramidKV & 34.80 & 14.00 & 13.20 & 50.80 & 15.00 & 17.36 & 25.93 & 45.00 & 40.00 & 59.17 & 75.17 & 35.49 \\
& SnapKV & 38.40 & 15.20 & 18.80 & 60.00 & 19.80 & \textbf{19.90} & 26.55 & 51.00 & \textbf{45.33} & \textbf{59.83} & 82.17 & 39.73 \\
& DapQ & \textbf{41.40} & \textbf{25.40} & \textbf{27.60} & 67.40 & \textbf{21.00} & 17.64 & \textbf{26.55} & \textbf{51.33} & 47.00 & 59.00 & \textbf{85.17} & \textbf{42.68} \\
\bottomrule
\end{tabular}
}
\label{table:helmet}
\end{table*}




\begin{table*}[t]
\centering
\small
\scriptsize 

\caption{Performance comparison of different methods across various LLMs on Needle-in-a-Haystack.}
\renewcommand{\arraystretch}{0.9}  

\label{table:needle}

\begin{tabular}{@{}>{\centering\arraybackslash}p{3.5cm} c l >{\centering\arraybackslash}p{1cm}@{}}  
\toprule
\textbf{LLM} & \textbf{KV Cache Size} & \textbf{Method} & \textbf{Acc} \\ \midrule
\multirow{13}{*}{Llama3-8B-Instruct} & & \textbf{FullKV} & \textbf{100.00} \\ \cmidrule(l){2-4} 
 & \multirow{4}{*}{256} & H2O       & 66.81 \\
 &                      & PyramidKV & 93.94 \\
 &                      & SnapKV    & 90.97 \\
 &                      & \textbf{DapQ} & \textbf{99.46} \\ \cmidrule(l){2-4} 
 & \multirow{4}{*}{128} & H2O       & 50.92 \\
 &                      & PyramidKV & 79.67 \\
 &                      & SnapKV    & 74.67 \\
 &                      & \textbf{DapQ} & \textbf{95.75} \\ \cmidrule(l){2-4} 
 & \multirow{4}{*}{64}  & H2O       & 42.37 \\
 &                      & PyramidKV & 55.45 \\
 &                      & SnapKV    & 61.70 \\
 &                      & \textbf{DapQ} & \textbf{68.34} \\ \midrule
\multirow{13}{*}{Llama3.1-8B-Instruct} & & \textbf{FullKV} & \textbf{98.02} \\ \cmidrule(l){2-4} 
 & \multirow{4}{*}{256} & H2O       & 61.18 \\
 &                      & PyramidKV & 78.30 \\
 &                      & SnapKV    & 74.84 \\
 &                      & \textbf{DapQ} & \textbf{84.70} \\ \cmidrule(l){2-4} 
 & \multirow{4}{*}{128} & H2O       & 47.61 \\
 &                      & PyramidKV & 65.23 \\
 &                      & SnapKV    & 61.45 \\
 &                      & \textbf{DapQ} & \textbf{70.34} \\ \cmidrule(l){2-4} 
 & \multirow{4}{*}{64}  & H2O       & 40.36 \\
 &                      & PyramidKV & 52.25 \\
 &                      & SnapKV    & 56.50 \\
 &                      & \textbf{DapQ} & \textbf{62.20} \\ \midrule
\multirow{13}{*}{Qwen2.5-7B-Instruct} & & \textbf{FullKV} & \textbf{94.23} \\ \cmidrule(l){2-4} 
 & \multirow{4}{*}{256} & H2O       & 75.64 \\
 &                      & PyramidKV & 83.80 \\
 &                      & SnapKV    & 84.30 \\
 &                      & \textbf{DapQ} & \textbf{85.11} \\ \cmidrule(l){2-4} 
 & \multirow{4}{*}{128} & H2O       & 70.45 \\
 &                      & PyramidKV & 74.80 \\
 &                      & SnapKV    & 73.64 \\
 &                      & \textbf{DapQ} & \textbf{76.25} \\ \cmidrule(l){2-4} 
 & \multirow{4}{*}{64}  & H2O       & 63.70 \\
 &                      & PyramidKV & 56.11 \\
 &                      & SnapKV    & 72.84 \\
 &                      & \textbf{DapQ} & \textbf{75.75} \\ \midrule
\multirow{13}{*}{Qwen3-8B} & & \textbf{FullKV} & \textbf{96.52} \\ \cmidrule(l){2-4} 
 & \multirow{4}{*}{256} & H2O       & 74.55 \\
 &                      & PyramidKV & 88.50 \\
 &                      & SnapKV    & 90.41 \\
 &                      & \textbf{DapQ} & \textbf{91.73} \\ \cmidrule(l){2-4} 
 & \multirow{4}{*}{128} & H2O       & 67.50 \\
 &                      & PyramidKV & 72.36 \\
 &                      & SnapKV    & 75.39 \\
 &                      & \textbf{DapQ} & \textbf{77.89} \\ \cmidrule(l){2-4} 
 & \multirow{4}{*}{64}  & H2O       & \textbf{62.70} \\
 &                      & PyramidKV & 61.32 \\
 &                      & SnapKV    & 59.73 \\
 &                      & \textbf{DapQ} & 61.98 \\ \bottomrule
\end{tabular}
\end{table*}

\begin{figure*}[t]
  \centering
  
  \begin{subfigure}[b]{0.7\linewidth} 
    \centering
    \includegraphics[width=\linewidth]{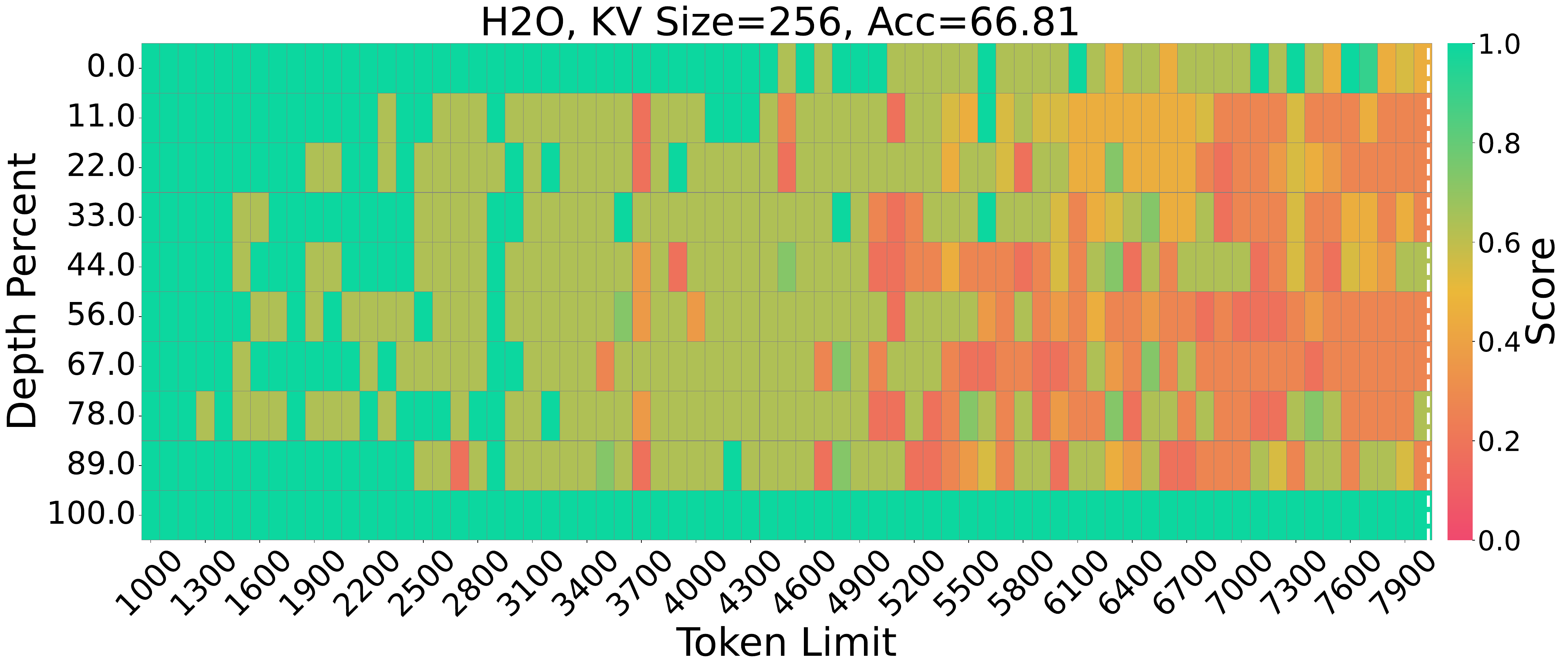}
  \end{subfigure}
  \par\vspace{0.2cm} 
  
  \begin{subfigure}[b]{0.7\linewidth}
    \centering
    \includegraphics[width=\linewidth]{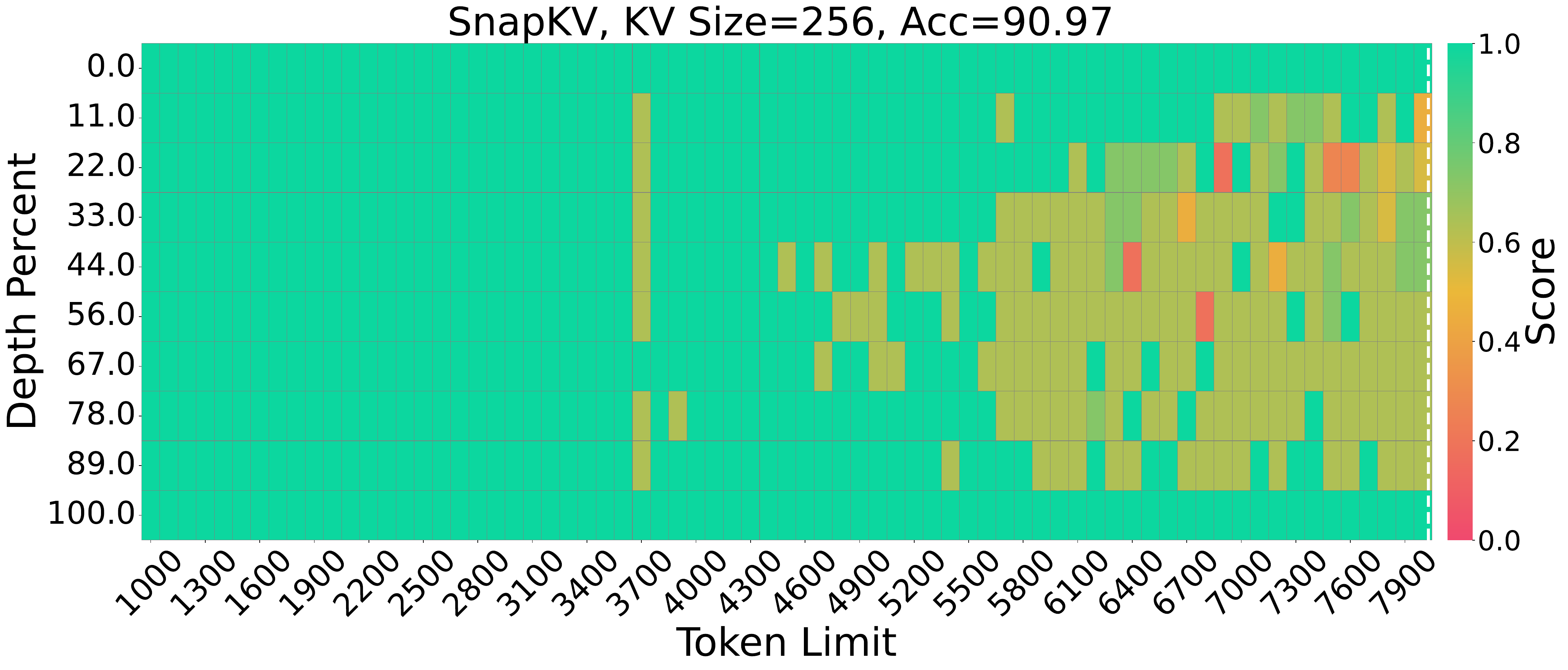}
  \end{subfigure}
  \par\vspace{0.2cm}
  
  \begin{subfigure}[b]{0.7\linewidth}
    \centering
    \includegraphics[width=\linewidth]{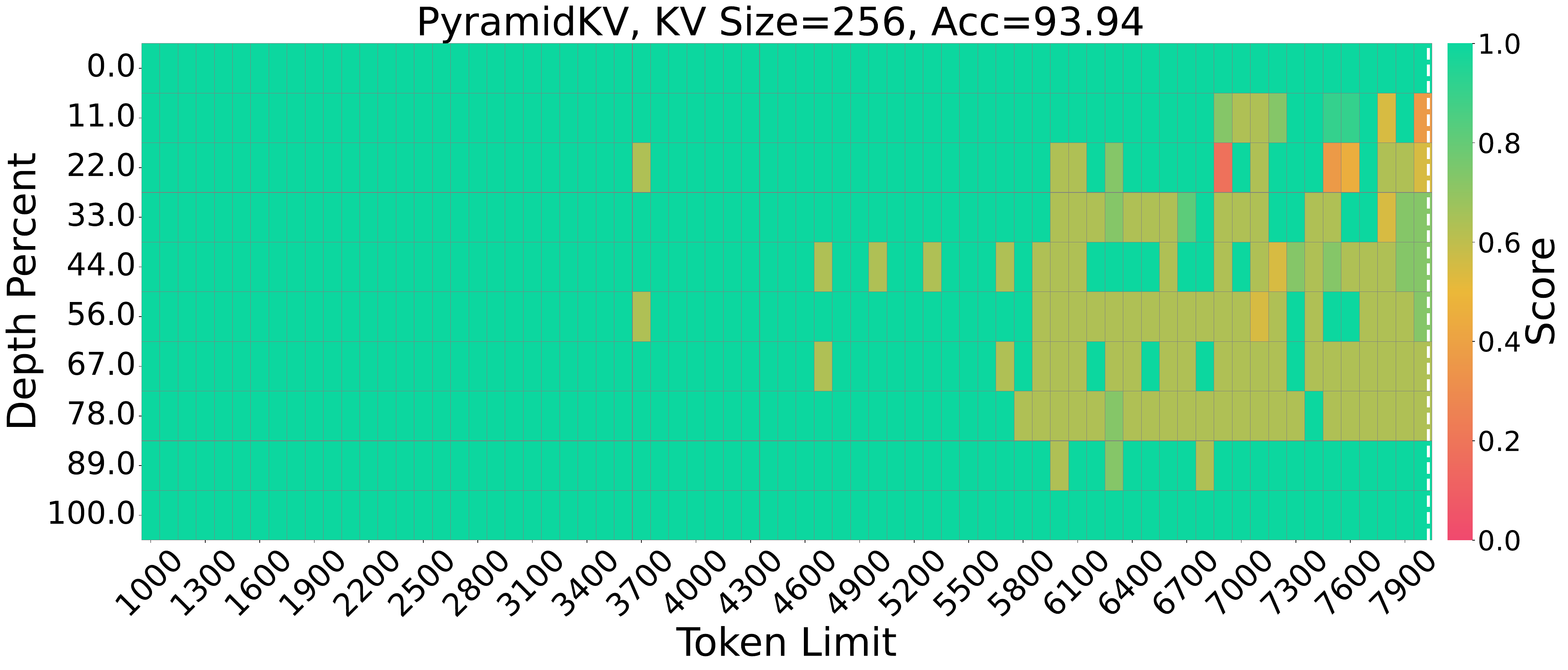}
  \end{subfigure}
  \par\vspace{0.2cm}
  
  \begin{subfigure}[b]{0.7\linewidth}
    \centering
    \includegraphics[width=\linewidth]{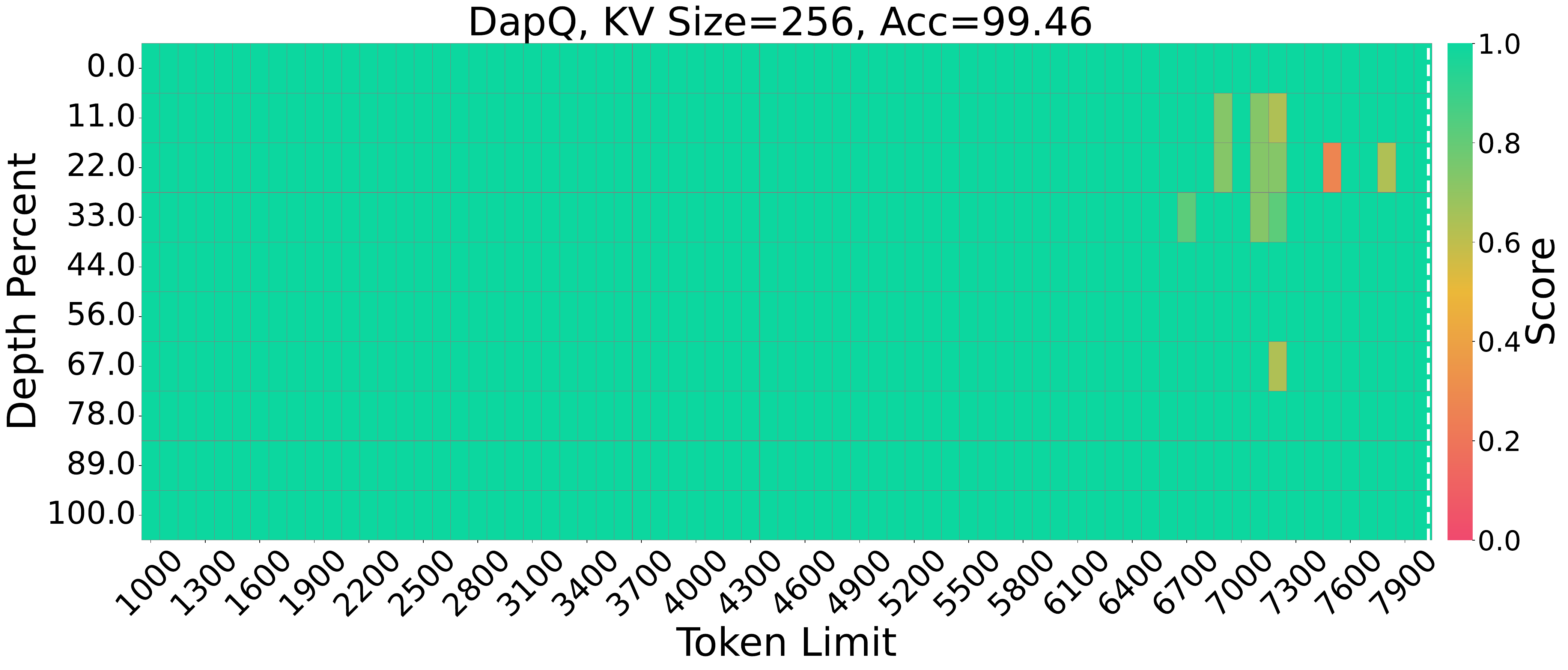}
  \end{subfigure}
  
\caption{Visualization of Needle-in-a-Haystack results. We take LLaMA-3-8B-Instruct (8k context, 256 KV size) as a representative example to demonstrate the performance differences. The vertical axis represents the depth percentage, and the horizontal axis represents the token length.}
  \vspace{-14pt}
  \label{fig:needle}
\end{figure*}

\end{document}